\documentclass[12pt]{article}
\usepackage{amsmath,amsfonts}
\usepackage{algorithm}
\usepackage{algorithmicx}
\usepackage{algpseudocode}

\usepackage{amsmath,amsfonts,amssymb,amsthm}
\usepackage{array}
\usepackage[caption=false,font=tiny,labelfont=sf,textfont=sf]{subfig}
\usepackage{textcomp}
\usepackage{stfloats}
\usepackage{url}
\usepackage{verbatim}
\usepackage{graphicx}
\usepackage{pgffor}%
\usepackage{natbib}
\usepackage{stmaryrd}
\usepackage{booktabs}
\usepackage{color}

\makeatletter
\bibpunct{(}{)}{;}{autheryear}{,}{;}
\allowdisplaybreaks

\allowdisplaybreaks
\@ifundefined{date}{}{\date{}}

\let\svthefootnote\thefootnote
\newcommand\freefootnote[1]{%
  \let\thefootnote\relax%
  \footnotetext{#1}%
  \let\thefootnote\svthefootnote%
}

\usepackage{xr}
\begin{document}
\def\spacingset#1{\renewcommand{\baselinestretch}%
{#1}\small\normalsize} \spacingset{1}
  \title{Enhancing convolutional neural network generalizability via low-rank weight approximation}
  \author{Chenyin Gao\thanks{ Department of Statistics, North Carolina State University, North Carolina
27695, U.S.A. Email: cgao6@ncsu.edu.}\hspace{.2cm},~ Shu Yang\thanks{
   Department of Statistics, North Carolina State University, North Carolina
27695, U.S.A. Email: syang24@ncsu.edu.}
~ and Anru R. Zhang\thanks{Department of Biostatistics \& Bioinformatics and Department of Computer Science, Duke University, North Carolina 27710, U.S.A. Email: anru.zhang@duke.edu}}

\maketitle
\bigskip
\begin{abstract}
Noise is ubiquitous during image acquisition. Sufficient denoising is often an important first step for image processing. In recent decades, deep neural networks (DNNs) have been widely used for image denoising. Most DNN-based image denoising methods require a large-scale dataset or focus on supervised settings, in which single/pairs of clean images or a set of noisy images are required. This poses a significant burden on the image acquisition process. Moreover, denoisers trained on datasets of limited scale may incur over-fitting. To mitigate these issues, we introduce a new self-supervised framework for image denoising based on the Tucker low-rank tensor approximation. With the proposed design, we are able to characterize our denoiser with fewer parameters and train it based on a single image, which considerably improves the model's generalizability and reduces the cost of data acquisition. Extensive experiments on both synthetic and real-world noisy images have been conducted. Empirical results show that our proposed method outperforms existing non-learning-based methods (e.g., low-pass filter, non-local mean), single-image unsupervised denoisers (e.g., DIP, NN+BM3D) evaluated on both in-sample and out-sample datasets. The proposed method even achieves comparable performances with some supervised methods (e.g., DnCNN). \freefootnote{This project was initiated while C. Gao participated in the 2021 HIV/AIDS summer camp at Duke University School of Medicine mentored by A. R. Zhang.} 
\end{abstract}

\noindent%
\vfill

\newpage
\spacingset{1.5} 


\section{Introduction}\label{sec:intro}
Noise in images, attributed to various factors such as noise corruption \citep{zhang2017image} and resolution limit \citep{buban2010high}, imposes great challenges in image processing \citep{buades2005review}. The removal of noise, i.e., denoising, is often a crucial step in advance of further tasks, e.g., image segmentation, recognition, and  classification. Among many image denoising methods in the literature (see, e.g., \citet{fan2019brief}, for a survey), deep learning frameworks, especially the convolutional neural networks (CNNs), stand out as a prominent approach. However, most of these frameworks focus on supervised settings, in which single/pairs clean images, e.g., \citep{zhang2017beyond, zhang2018ffdnet} or a set of noisy images, e.g., \citep{lehtinen2018noise2noise, krull2019noise2void} are required for learning the denoising mapping $f_{\boldsymbol{\theta}}(\cdot)$. But collecting a large number of useful images poses a burden to time and budgets, and sometimes such scrutinized images might not even exist in practice. In addition, leveraging millions of image datasets almost exclusively requires Graphics Processing Units (GPUs), which presents another challenge to run on conventional desktops or specialized computer hardware.

Recently, various denoising methods were proposed that only take single images of interest as input to the deep learning process. \cite{ulyanov2018deep} proposed a single-image deep learning model for image recovery using image priors. \cite{soltanayev2018training} proposed to train the denoisers on single noisy images using Stein's unbiased risk estimator to deal with Gaussian additive noise, with extensions in \cite{cha2019fully, zhussip2019extending}. However, these methods require that the noise levels be known a priori. \cite{wu2020unpaired} introduced the dilated blind-spot network to explicitly estimate the noise levels from unpaired noisy images for better performance. \cite{quan2020self2self} developed a Self2Self dropout scheme for single-image unsupervised learning, which achieves much better denoising performance than the existing self-similarity non-local methods. \cite{zheng2020unsupervised} proposed another competitive single-image denoiser that can handle more complex noise distributions by fusing deep learning-based methods with traditional non-learning-based Gaussian denoisers (e.g., BM3D, NLM) within the plug-and-play framework. In theory, the convergence of exploiting plug-and-play priors for fusing two bounded denoisers has been established in \cite{shi2023provable, shi2023regularization}. More recently, \cite{huang2021neighbor2neighbor} trained image denoisers using single noisy images by generating training pairs with a random neighbor sub-sampler. This approach avoids heavy dependence on assumptions about the noise distribution. However, those single-image convolutional neural networks often lack the ability for generalization. Since our primary interest is in self-supervised learning methods, it is unlikely they will be exposed to sufficient noise types when implemented in practice. Such iterative algorithms tend to be trapped in a sharp loss surface, which bears the risk of producing a highly unstable prediction. As it is needed, more generalizability of neural networks should be considered. 

A number of recent works have investigated the generalizability of deep neural networks and the potential methods for improving it. Apart from image denoising, there have been many works focusing on alleviating over-parameterization in order to improve the generalizability of the neural networks, such as parameter pruning \citep{han2015learning}, weights sharing \citep{chen2015compressing}, weights tensorization \citep{huang2017highly}, weights binarization \citep{hubara2016binarized}, knowledge distillation \citep{hinton2015distilling}, etc. Moving forward, some researchers applied the mentioned methods together to achieve a simpler model representation \citep{han2015deep,kozyrskiy2020cnn}. Nonetheless, most of them are subject to model accuracy decrease when a simplified model structure is deployed, even if a fine-tuning process is included afterward. Thus, it still remains a challenging task to remove the redundant parameters while maintaining model performance, especially in few-shot or one-shot learning tasks (e.g., single-image denoising).

Meanwhile, tensor decomposition provides an insightful perspective for investigating the generalizability of neural networks. One recent work of \citet{arora2018stronger} characterized the generalizability of a neural network based on the canonical polyadic decomposition (CPD) (see \citet{kolda2009tensor} for a survey on tensor decomposition), and achieved a provable generalization error bound. Hereafter, the low-rankness of weight kernels has been considered as a critical measuring tool when the number of parameters, i.e., generalization ability, is of interest. Most current applications in the image restoration paradigms are based on the higher-order tensor decomposition techniques (e.g., CP decomposition \citep{wu2018fused},  tensor train decomposition \citep{novikov2015tensorizing,phan2020tensor}, tensor SVD \citep{zhang2020denoising}) to exploit the nature of latent low-rankness structures. Although exact tensor factorization typically tends to be computationally intractable, the optimal solution can be obtained via a ``warm-start" based on a proper tensor matricization \citep{richard2014statistical}. However, such vanilla tensor decompositions, even paired with a ``warm-start" can be numerically unstable and may not converge to the desired optimum in the end \citep{rabanser2017introduction}.  In addition, it is worth noting that some weight kernels among the practical deep learning networks (e.g., the well-trained VGG-16 and WRN-28-10) do not necessarily admit the low-rank CP representation as demonstrated by  \citet{li2020understanding}.

\paragraph{Main contributions} 
To overcome the above limitations, this paper proposes a new algorithm that can boost the model performance with the compressed parameters. We focus on low-rank approximation  \citep{lebedev2014speeding,kim2015compression}, primarily for its clean format and good interpretability. Owing to the nature that the weight kernels in the CNN are usually in high-order (i.e., 4D tensors to be explained later), we specifically resort to Tucker decomposition \citep{tucker1966some} to exploit the low rankness with the aid of variational Bayesian matrix factorization (VBMF) for rank selection. The Tucker decomposition is particularly desirable in our denoising framework since it allows low-rank approximation on a part of tensor modes. Such tensor decomposition has also been exploited to realize the CNN compression schemes in other scenarios (see \citet{kim2015compression,nakajima2011global,bulat2019matrix} and the references therein). 
In addition, we adopt the newly proposed Tucker decomposition-based weight distortion scheme \citep{lee2019learning} within our single-image denoising context, in which the original model structure is retained and only the weight values are twisted by low-rank approximation after certain times of iterations. Then, such a weight distortion method is incorporated to the similar alternating direction method of multipliers (ADMM) framework as \citet{zheng2020unsupervised} to increase its generalization ability. The main contributions of this paper are summarized as follows:
\begin{enumerate}
    \item We apply the low-rank tensor decomposition to adjust the direction of ADMM to enable denoising on single images and reduce the computational complexity. By applying the Tucker low-rank approximation to the weight tensors at each twist stage, we can alleviate stochastic variance induced by the stochastic gradient descent optimizer while preventing the likelihood from trapping in undesirable local minima (to be explained in Section \ref{sec:model-optimization}).
    \item  The hyperparameter ranks are selected by the data-driven VBMF scheme to bypass any intermittent interventions. The components of the weight tensor with small variation are pruned via VBMF on mode-3 and mode-4, and the resulting refined neural networks can achieve a more robust estimation while preserving the original training (in-sample) performance.
    \item 
    We also conduct extensive numerical experiments on both the benchmark and real-world noisy image datasets that show our method significantly outperforms the existing methods in the literature. It is worth noting that our method can even achieve a higher peak signal-to-noise ratio (PSNR) under many scenarios. Moreover, our proposed framework can also achieve significantly better image denoising performance on other external (out-sample) testing datasets, indicating its good generalization ability. Our implementation code and experimental details will be made publicly available after the acceptance of this manuscript.
\end{enumerate}
 
\section{Preliminaries and Notation}\label{sec:prelim}

Lowercase letters (e.g., $x,y$),  bold lower case letters (e.g., $\boldsymbol{x}, \boldsymbol{y}$), and uppercase letters (e.g., $X, Y$) are used to denote scalars, vectors, and matrices, respectively, unless defined otherwise. Let $X_{ij}$, $X_{i.}$ and $X_{.j}$ be the $(i,j)$th entry, the $i$th row, and the $j$th column of $X$, respectively. We use calligraphic-style letters $\mathcal{X}$ and $\mathcal{Y}$ to denote higher-order tensors with their entries $\mathcal{X}_{ijk}$ and $\mathcal{Y}_{ijk}$, respectively. For a general tensor $\mathcal{X}\in \mathbb{R}^{p_1\times p_2\times \cdots \times p_n}$, its mode-$1$ tensor-matrix product with a matrix $W\in \mathbb{R}^{r_1\times p_1}$ is
\begin{align*}
&\mathcal{X}\times_1
W \in \mathbb{R}^{r_1\times p_2 \times \cdots \times p_n},\\
&(\mathcal{X}\times_1
W)_{i_1,\cdots,i_n}=
\sum_{j=1}^{p_1} 
\mathcal{X}_{j,\cdots,i_n}
W_{i_1,j}.
\end{align*}
The general mode-$k$ tensor-matrix product is defined similarly. The mode-$k$ matricization of $\mathcal{X}$ is obtained by unfolding it to the matrix $\mathcal{X}_{(k)}$ along the $k$th mode, and the multilinear rank (Tucker rank) is then defined as $(\text{rank}(\mathcal{X}_{(1)}), \cdots, \text{rank}(\mathcal{X}_{(n)}))$. The Tucker decomposition of an order-$n$ multilinear rank-$(r_1,r_2,\cdots, r_n)$ tensor $\mathcal{X}\in \mathbb{R}^{p_1\times p_2\times \cdots \times p_n}$ is
\[\mathcal{X}=\mathcal{S}\times_1 U_1 \cdots \times_{n}U_n=\llbracket\mathcal{S};U_1,U_2,\cdots,U_n\rrbracket,\]
where $\mathcal{S}\in \mathbb{R}^{r_1\times r_2\times \cdots\times r_n}$ is a core tensor, and $U_i\in \mathbb{R}^{p_i\times r_i}, i=1,\cdots, n$ are the factor matrices. Higher-order orthogonal iteration (HOOI) \citep{de2000best,zhang2018tensor} is a classic algorithm that finds the rank-$(r_1,r_2,\cdots, r_n)$ approximation for $\mathcal{X}$ and we denote the result from HOOI as $\Tilde{\mathcal{X}}$. In addition, the Frobenius norm of a matrix $X$ and a tensor $\mathcal{X}$ is denoted as $\|X\|_F=\sqrt{\sum_{i_1,i_2} X_{i_1,i_2}^2}$ and $\|\mathcal{X}\|_F=\sqrt{\sum_{i_1,\cdots,i_n} \mathcal{X}_{i_1,\cdots,i_n}^2}$, respectively.

The standard 2D convolutional operation of a CNN can be presented in a tensor format: with $\mathcal{X}\in \mathbb{R}^{h\times w\times s}$ as the input, the convolutional operation generates a new tensor $\mathcal{Y}\in \mathbb{R}^{\widehat{h}\times \widehat{w}\times \widehat{s}}$ based on an order-4 weight kernel tensor $\mathcal{W}\in \mathbb{R}^{d\times d\times s \times \widehat{s}}$. Elementwisely, we have 
\begin{equation}
\label{eq:conv2D}
    \mathcal{Y}_{\widehat{h}\widehat{w}\widehat{s}}=
\sum_{i=1}^d\sum_{j=1}^d \sum_{k=1}^s
\mathcal{W}_{ijk\widehat{s}}\times \mathcal{X}_{h_i w_j k},
\end{equation}
where $d$ is the spatial width of the weight kernel $\mathcal{W}$, $h_i=(\widehat{h}-1)\delta + i -p$, $w_j=(\widehat{w}-1)\delta+j-p$, $\delta$  is the stride size,  and $p$ is the zero-padding size.

\section{The Proposed Methods}\label{sec:method}

Our denoising algorithm starts with the following additive model \citep{buades2005non,aharon2006k, dabov2007image, gu2014weighted}:
\begin{equation}\label{eq:noisy}
    \mathcal{Y}=\mathcal{X}^*+\mathcal{E},
\end{equation}
where $\mathcal{Y}\in \mathbb{R}^{h\times w\times s}$ is the noisy image, $\mathcal{X}^*\in \mathbb{R}^{h\times w\times s}$ is the latent clean image, and $\mathcal{E}\in \mathbb{R}^{h\times w\times s}$ is the noise, each  with height $h$, width $w$, and number of channels $s$.
Below, we introduce the proposed algorithm to recover $\mathcal{X}^*$ step by step. The pseudocode of the overall procedure is summarized in Algorithm \ref{al:U-net+BM3D+T}. The sub-algorithms Variational Bayesian matrix factorization (VBMF) \citep{nakajima2013global} and Partial high-order orthogonal iteration (PHOOI) \citep{de2000best} are described in the supplementary materials.

\subsection{Model Formulation}\label{sec:model-formulation}

\begin{algorithm*}[!t]
\caption{Proposed Optimization Scheme}
   \label{al:U-net+BM3D+T}
    \small
\begin{algorithmic}
   \State {\bfseries Input:} Single noisy image $\mathcal{Y}$ 
   \State {\bfseries Initialization:} $\theta^{(0)}$ by \texttt{PyTorch}, $\mathcal{M}^{(0)}=\mathcal{Y}, \mathcal{A}^{(0)}=\mathbf{0}_{h\times w\times s},  \rho=10^{2}, \eta=.5 ,\mathcal{S}=\{S_D,2S_D,\cdots\}, S_D=200$\;
   \For{$k=0$ {\bfseries to} $K$}
   \For{$i=0,1,\cdots, m$ \hfill \Comment{ train by $m$ mini-batches}\\}
   \State $\theta^{(k), i+1}=
  \arg\min_\theta\left\{
  {2\sigma^{-2}}\|\mathcal{Y}-f_{\theta^{(k),i}}^{(k)}(\mathcal{Y})\|_F^2+
  {\rho}
  \|f_{\theta^{(k),i}}^{(k)}(\mathcal{Y})+\mathcal{A}^{(k)}-\mathcal{M}^{(k)}\|^2_F/{2}
  \right\}$
  \If{$km+i\in \mathcal{S}$}
  
  \For{$\mathcal{W}^{(k)}_n\in \boldsymbol{\theta}^{(k),i+1}$ and $n\neq 1$  \hfill\Comment{ skip the first layer}\\}
  \State $(r_3,r_4) = \text{VBMF}(\mathcal{W}_n^{(k)},\sigma^2)$
  \State $(\mathcal{G}^{(k)}, U_{r_3}^{(3)}, U_{r_4}^{(4)})=\text{PHOOI}(\mathcal{W}_n^{(k)},r_3,r_4)$
  \State $\Tilde{\mathcal{W}}_n^{(k)}=\mathcal{G}^{(k)}\times_3 {U}^{(3)}_{r_3}\times_4 {U}^{(4)}_{r_4}, \mathcal{W}_n^{(k)}=
  \Tilde{\mathcal{W}}_n^{(k)}.
 $
  \EndFor
  \EndIf
   \EndFor
   \State $\theta^{(k+1)} =\theta^{(k),m},f^{(k+1)}=
  f_{\theta^{(k+1)}}, \mathcal{X}^{(k+1)} = f^{(k+1)}(\mathcal{Y})$
  \State $\mathcal{M}^{(k+1)}= \arg\min_{\mathcal{M}}\left\{ 
  R(\mathcal{M}^{(k)})+
  {\rho}
  \|
  f^{(k+1)}(\mathcal{Y})-\mathcal{M}^{(k)}+
  \mathcal{A}^{(k)}
  \|_F^2/{2}\right\}$
  \hfill\Comment{ $\lambda$ is absorbed in $\rho$}
  \State $\mathcal{A}^{(k
  +1)}=
  \mathcal{A}^k+
  \eta\left\{f^{(k+1)}(\mathcal{Y})-\mathcal{M}^{(k+1)}\right\} $
   \EndFor
  \State {\bfseries Output:} $\widehat{\mathcal{X}}=f^{(K)}(\mathcal{Y})$
\end{algorithmic}
\end{algorithm*}

To construct an initial objective function, we introduce the {\it maximum a posterior} (MAP) framework, which aims at maximizing the posterior distribution $p(\mathcal{X}\mid \mathcal{Y})$ as follows:
 \begin{equation}\label{eq:obj0}
    \max_\mathcal{X} \ln p(\mathcal{X}\mid \mathcal{Y})
    \propto
    \max_\mathcal{X}\{ \ln p(\mathcal{Y}\mid \mathcal{X})+
    \ln p(\mathcal{X})\}. \end{equation}
    Here, $p(\mathcal{X})$ is the prior distribution of $\mathcal{X}$, representing the information before acquiring the image, and  
    $p(\mathcal{Y}\mid \mathcal{X})$  is the likelihood function. For example, if  
    $p(\mathcal{X})\propto \exp\{-\lambda R(\mathcal{X})\}$ and
    $\mathcal{Y}\mid \mathcal{X} \sim \mathcal{N}(\mathcal{X}, \sigma^2\mathbf{I})$,  
    \eqref{eq:obj0} becomes a loss minimization problem 
    \begin{equation}\label{eq:obj}
      \min_\mathcal{X}l(\mathcal{Y},\mathcal{X})= \min_\mathcal{X}
       \left\{
       \frac{1}{2\sigma^2}\|\mathcal{Y}-\mathcal{X}\|^2_F+\lambda R(\mathcal{X})
       \right\},
    \end{equation}
    where $R(\mathcal{X})$ can be seen as a regularization term. In practice, the loss function in \eqref{eq:obj} is high-dimensional and often non-convex, making direct computation unstable. To resolve the computational challenge, we introduce $\mathcal{M}$, $\mathcal{A}\in \mathbb{R}^{h\times w \times s}$ and rewrite the loss function of \eqref{eq:obj} in an augmented Lagrangian format:
    \begin{gather}
     l(\mathcal{X},\mathcal{Y},\mathcal{M},\mathcal{A};\rho,\lambda) = \frac{1}{2\sigma^2}
    \|\mathcal{Y}-\mathcal{X}\|_F^2 + \lambda R(\mathcal{M})+
    \frac{\rho}{2}
    (\|\mathcal{X}-\mathcal{M}+\mathcal{A}\|_F^2-\|\mathcal{A}\|_F^2),
    \label{eq:augmented-loss}
    \end{gather}
    where $\lambda$ and $\rho$ are two hyperparameters to be further discussed in Section \ref{sec:exp}. By formulating \eqref{eq:obj} into  \eqref{eq:augmented-loss}, the constraints are split into two parts, $\|\mathcal{Y}-\mathcal{X}\|_F^2$ and $R(\mathcal{M})$. As a result, we only need to minimize the distance between $\mathcal{X}$ and $\mathcal{Y}$ while controlling the complexity of the auxiliary variable $\mathcal{M}$.
    To this end, we introduce a dual variable $\mathcal{A}$ to control the proximity between $\mathcal{M}$ and $\mathcal{X}$ to promote the constraint $\mathcal{X}-\mathcal{M}=0$.

\subsection{Computational Method}\label{sec:model-optimization}
\begin{figure}[htbp]
       \centering
       \includegraphics[width=.8\linewidth]{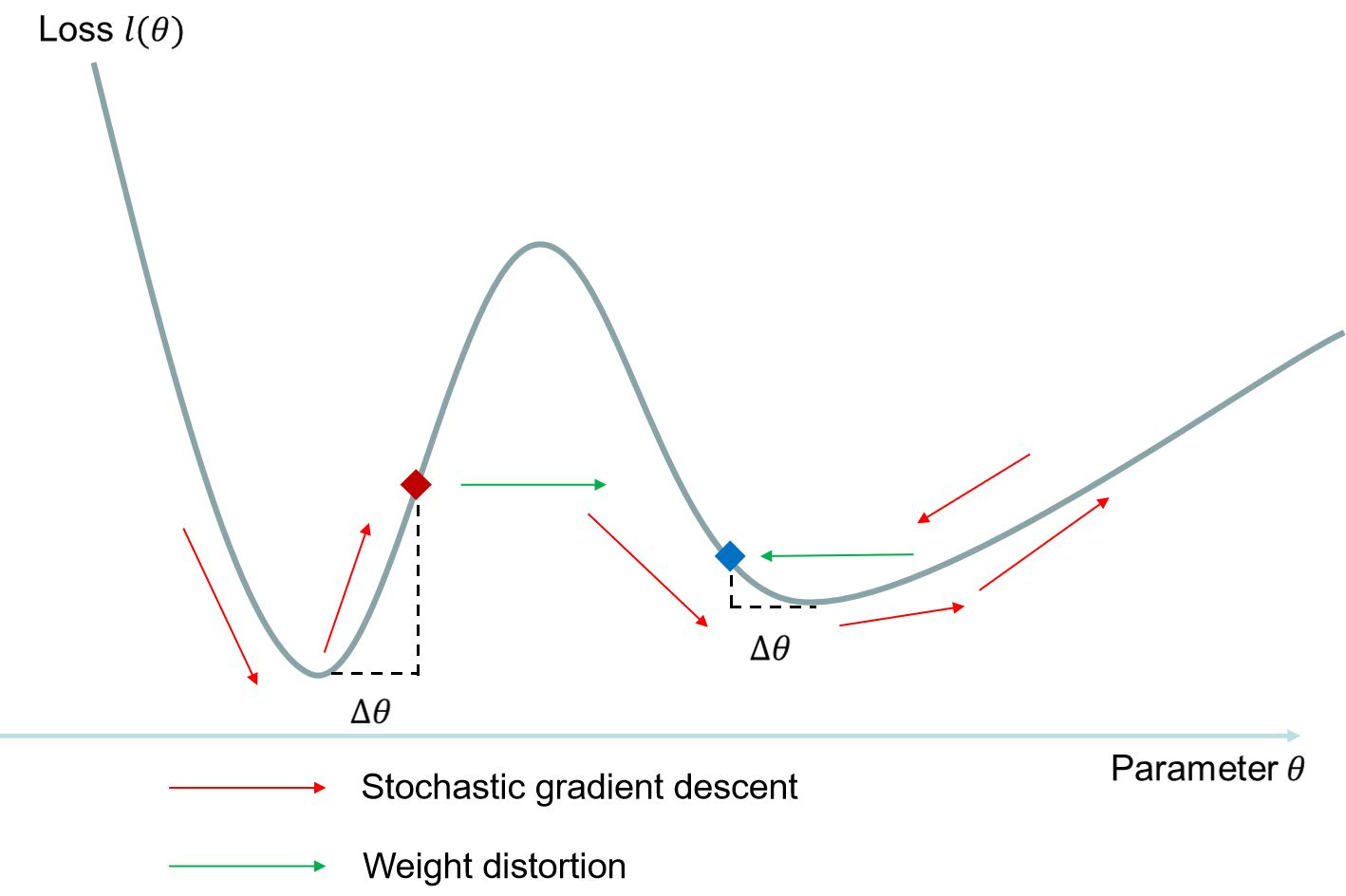}
       \caption{A schematic illustration of weight distortion in the stochastic gradient descent training process}
       \label{fig:1}
   \end{figure}

We solve \eqref{eq:augmented-loss} by alternating direction method of multipliers (ADMM) \citep{boyd2011distributed}, which updates the auxiliary and dual variables alternatively and iteratively. The procedure includes the following three sub-problems.
    \begin{enumerate}
        \item Update $\mathcal{X}$: 
        We establish a CNN-based mapping $f_{\boldsymbol{\theta}}(\cdot)$ from the noisy image $\mathcal{Y}$ to the update of $\mathcal{X}$. The mapping $f_{\boldsymbol{\theta}}(\cdot)$ essentially characterizes the features of $\mathcal{X}$ parametrized by $\boldsymbol{\theta}$, where $\boldsymbol{\theta}$ is the collection of trainable parameters, including a convolution weight tensor in each layer. We note $\mathcal{W}_n\in \mathbb{R}^{d\times d\times s\times \widehat{s}}$ as the convolution weight tensor for the $n$th layer. After substituting $\mathcal{X}$ with $f_{\boldsymbol{\theta}}(\mathcal{Y})$,  $\boldsymbol{\theta}$ can be iteratively updated through the backpropagation (BP) scheme \citep{rumelhart1986learning}. For example, in the $(k+1)$th iteration, we update the parameters by \begin{equation}\label{eq:sub1} \begin{split} &\boldsymbol{\theta}^{(k+1)}=
    \arg\min_{\boldsymbol{\theta}}
    l\{f_{\boldsymbol{\theta}}(\mathcal{Y}),\mathcal{Y},\mathcal{M}^{(k)},\mathcal{A}^{(k)};\rho,\lambda\},\\
    &\mathcal{X}^{(k+1)}=f_{\boldsymbol{\theta^{(k+1)}}}(\mathcal{Y}),
    \end{split}
    \end{equation} 
    where the subscript $k$ or $k+1$ denotes the number of epochs. When our target image has a large size (e.g., high-resolution micrographs), we can apply mini-batch stochastic gradient descent on image patches to accelerate the computation.
        \item Update $\mathcal{M}$: We update  $\mathcal{M}$ by applying a proximal mapping $\Upsilon(\cdot)$. This sub-problem is equivalent to applying a Plug-and-Play technique to $\mathcal{X}^{(k+1)}+\mathcal{A}^{(k+1)}$, which can be generalized to any existing denoisers, e.g., BM3D \citep{dabov2007image} and NLM \citep{buades2005non} among others \citep{venkatakrishnan2013plug,zheng2020unsupervised}. Specifically,     
        \begin{equation}
           \label{eq:sub2}
           \begin{split}
           \mathcal{M}^{(k+1)}=& \arg\min_{\mathcal{M}}
     \Big\{R(\mathcal{M})+\frac{\rho}{2}\|\mathcal{X}^{(k+1)}-\mathcal{M}+\mathcal{A}^{(k+1)}\|_F^2\Big\}\\
     =&
     \Upsilon(\mathcal{X}^{(k+1)}+\mathcal{A}^{(k+1)}).
     \end{split}
     \end{equation}
        \item Update $\mathcal{A}$: As the variable $\mathcal{A}$ only appears in the last term of our augmented loss function, it can be updated by
        \begin{equation}
            \label{eq:sub3}
            \mathcal{A}^{(k+1)}=\mathcal{A}^{(k)}+
            \eta (f_{\boldsymbol{\theta}^{(k+1)}}(\mathcal{Y})-\mathcal{M}^{(k+1)}),
        \end{equation}
        where $\eta$ can be seen as the learning rate of ADMM.
    \end{enumerate}
In Algorithm \ref{al:U-net+BM3D+T}, a Tucker low-rank approximation distortion is applied to the weight tensors in all CNN layers except the first one for the update of $\mathcal{X}$ at the pre-specified twist steps, i.e., $S_D, 2S_D, \cdots$, while retaining the original structures of the approximated layers. This strategy not only boosts the model's generalizability but also obviates significant damage to model performance.
    Specifically, the model generalizability corresponds to the flatness of the loss function \eqref{eq:obj} near the local minimum \citep{baldassi2020shaping}. If the loss surface is steep, measured by $\partial l/\partial \boldsymbol{\theta}$, small turbulence of $\boldsymbol{\theta}$ may lead to a deterioration of performance. Thus it is often more desirable to find another local minimum in which minor weight distortion only makes a slight impact on the value of training loss, i.e., $\partial l/\partial \boldsymbol{\theta}$ is of small magnitude. Specifically, our training procedure is coupled with such low-rank approximation after every $S_D$ 
   iterations to achieve stable convergence, where $S_D$ is a carefully selected hyperparameter that controls the trade-off between model quality and generability. This scheme can even boost the resulting model performance on various occasions. Figure \ref{fig:1} gives a schematic illustration: given the same level of estimation error due to the stochastic variability, represented by $\Delta\boldsymbol{\theta}$, the weight distortion scheme allows us to settle down on the blue point, which is more desirable than the red point when no weight adjustment is implemented. 
  In addition, through the low-rank approximation of the 4D weight tensor, the computation time at that step is significantly reduced. For example, the common kernel weight tensors $\mathcal{W}$ typically have small spatial mask size $d$ but large numbers of input/output channels $s$ and $\widehat{s}$. Thus, we apply Tucker decomposition with low-rank structures on modes-3 and 4:
   \begin{equation*}
   \begin{split} &\mathcal{W}=\mathcal{G}\times_3 {U}^{(3)}_{r_3}\times_4{U}^{(4)}_{r_4},\\
   & \text{with }
   \mathcal{G}\in \mathbb{R}^{d\times d \times r_3\times r_4},
   {U}^{(3)}_{r_3}\in \mathbb{R}^{s\times r_3},
   {U}^{(4)}_{r_4}\in \mathbb{R}^{\widehat{s}\times r_4},
   \end{split}
   \end{equation*}
   where $\mathcal{G}$ is the core tensor with loading matrices ${U}^{(3)}_{r_3}$ and ${U}^{(4)}_{r_4}$ for mode-3 and mode-4, respectively. Then the original 2D convolution operator in \eqref{eq:conv2D} can be rewritten in three consecutive convolutions as follows:
   \begin{align*}
       &\mathcal{Y}^{(1)}_{hw r_3}=
       \sum_{i=1}^{s} U_{r_3,ir_3}^{(3)}\mathcal{X}_{hwi},\\
       &\mathcal{Y}^{(2)}_{\widehat{h}\widehat{w}r_4}=
       \sum_{i=1}^d\sum_{j=1}^d \sum_{k=1}^{r_3}
\mathcal{G}_{ijkr_4}\times \mathcal{Y}^{(1)}_{h_i w_j k},\\
&\mathcal{Y}_{\widehat{h}\widehat{w}\widehat{s}}=
\sum_{i=1}^{r_4} U_{r_4,\widehat{s}i}^{(4)}\mathcal{Y}^{(2)}_{\widehat{h}\widehat{w}i}.
   \end{align*}
   The compression ratio of computation time of this convolution operation is
\begin{equation}
    d^2s\widehat{s}/\left(d^2r_3r_4+ sr_3+\widehat{s}r_4\right),
    \label{eq:CR}
\end{equation}
which is typically greater than $1$ for small values of ranks $r_3$ and $r_4$. In summary, the weight tensors are approximated by the VBMF-aided Tucker decomposition to filter out any unnecessary redundancy at each twist stage, and the original 2D convolution operator can be replaced with three consecutive convolutions to speed up the gradient computation. Throughout our training, the original structures of the approximated layers are retained since only the numerical values of the weight tensors are substituted. By applying the Tucker low-rank approximation to the weight tensors at each twist stage, we can alleviate stochastic variance induced by the stochastic gradient descent optimizer while preventing the likelihood of trapping in undesirable local minima. In addition, the weight distortion step does not change the structure of the convolution structure, which avoids significant in-sample quality drop but generalizes better to out-sample datasets.
\subsection{Rank Selection}\label{sec:rank:selection}
The selection of optimal Tucker rank can be challenging in practice. One popular data-driven approach is cross-validation, which can be computational intensive to apply in  the deep learning framework. 
Other researchers suggest a human-in-the-loop approach for rank tuning, in which the ``elbow" points are sought to partition the spectral values of the matricized weight tensors. However, this approach relies on the manual intervention at each iteration and does not allow the framework to be end-to-end. Towards this reason, we deploy the variational Bayesian matrix factorization (VBMF) sub-algorithm into our proposed optimization scheme for choosing $r_3$ and $r_4$, which achieves global optimality under certain conditions \citep{nakajima2013global}. The detailed VBMF algorithm is illustrated in the supplementary materials.

\section{Experiments}\label{sec:exp}
\begin{figure*}[!t]
    \centering
    \includegraphics[width =.85\linewidth]{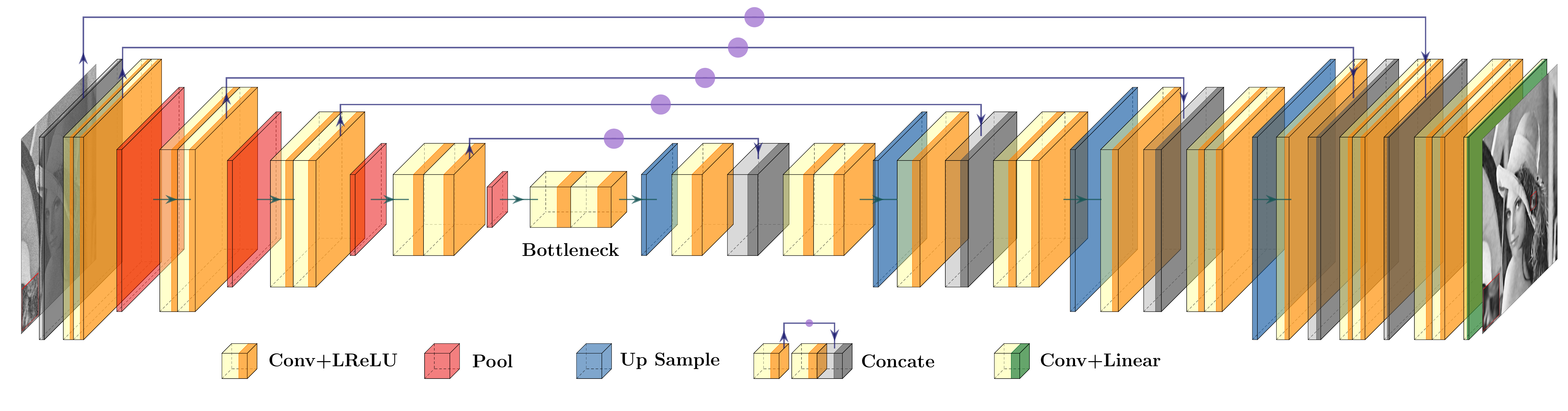}
    \caption{Illustration of the U-net deep learning framework. The left side (before the Bottleneck) is the encoder and the right side (after the Bottleneck) is the decoder.}
        \label{fig:u-net}
\end{figure*}

We assess the proposed single-image denoiser method on real-world noisy images, synthetic images corrupted by Gaussian noise and synthetic images corrupted by Poisson-Gaussian noise. For simplicity, all experiments are done on gray-scale images; colored image denoising can be done similarly. The deep learning network structure of $f_{\boldsymbol{\theta}}(\cdot)$ are chosen to be the U-net architecture with a 5-layer encoder and decoder \citep{ronneberger2015u} (see Figure \ref{fig:u-net} for a schematic plot); the denoiser $\Upsilon(\cdot)$ is chosen as the block-matching and 3D filtering (BM3D) \citep{dabov2007image}. The hyper-parameters $\rho$ and $\eta$ in the ADMM algorithm are set to be $100$ and $0.5$, respectively. The noise level $\sigma$ is estimated following the method in \citet{chen2015efficient}. Our algorithm is termed as NN+BM3D+T.

In the low-rank approximation step, we select the ranks $r_3, r_4$ by VBMF every $S_D=200$ iterations. We apply ADAM (a method for stochastic optimization) \citep{kingma2014adam} for mini-batch iterative training with the batch size $128$ and a decayed learning rate schedule: $0.01$ for the first $30$ epochs, $0.002$ for the next $30$ epochs, and $0.0004$ for the last 40 epochs. All networks are implemented in Pytorch with one standalone Intel(R) Core(TM) i7-8565U CPU, 16 RAM computer.

\subsection{Results}\label{sec:results}
\begin{figure*}[!t]
      \includegraphics[width=\linewidth]{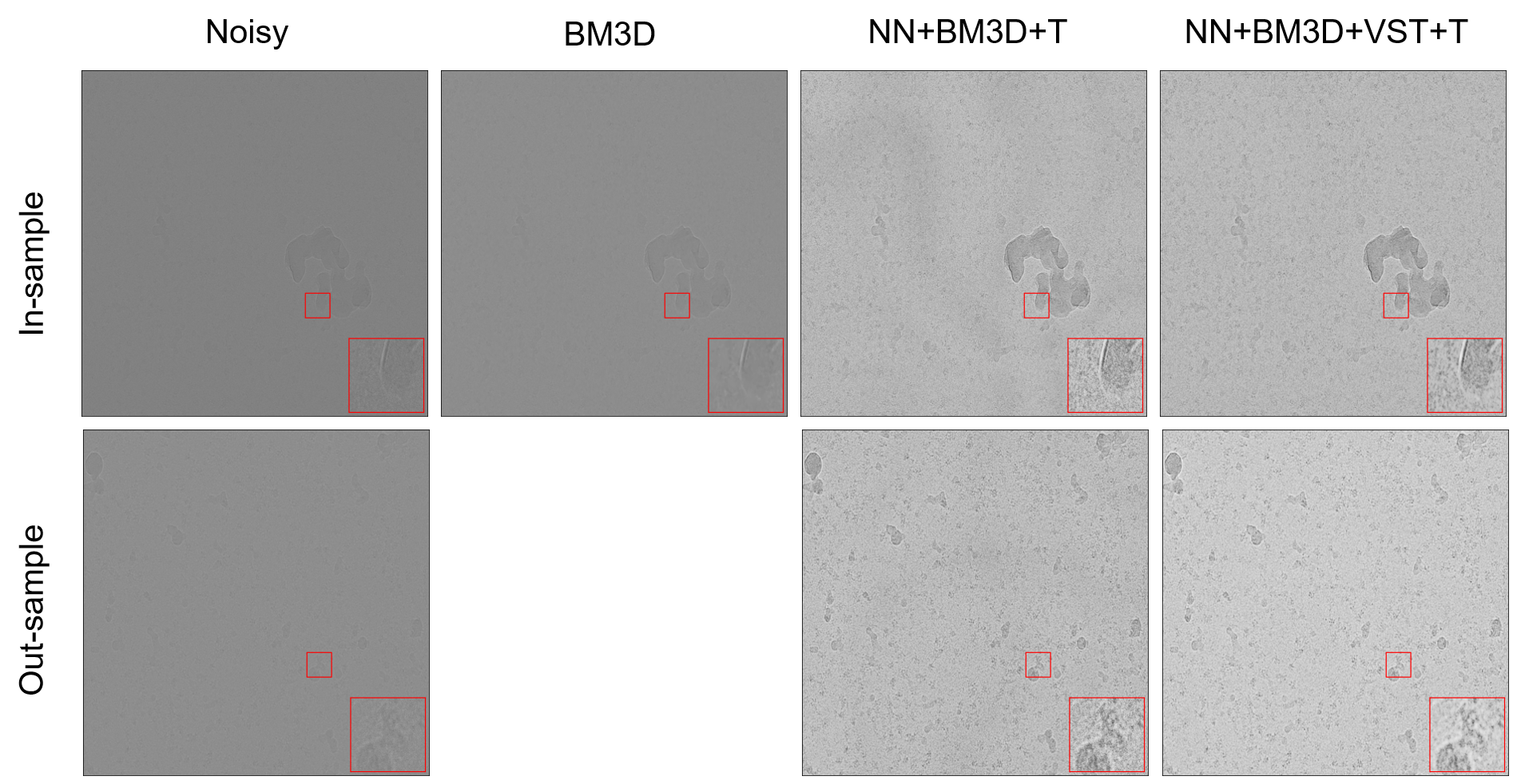}
    \caption{Visualized in-sample (top) and out-sample (bottom) performance comparison of on two raw micrographs of SARS-CoV-2 2P protein microscopy. The hyper-parameters for the ADMM framework is chosen as $\eta = 0.5, \rho = 100$. Recall BM3D is a non-learning-based method and therefore does not provide a out-sample denoiser.}
    \label{fig:real}
\end{figure*}

\paragraph{Experiments on real-world noise}
We first apply our proposed algorithm to a real-world dataset of the SARS-CoV-2 2P protein produced by cryo-electron microscopy (cryo-em). The image includes $5760\times 4092$ pixels with a pixel size around $1.058\mathring{\text{A}}$. The raw image is obtained by running the MotionCor2 algorithm in \citet{zheng2017motioncor2} on the collected movie frames, and its image quality is substantially limited by various factors, such as dose fractionation, speciman heterogeneity, and radiation damage \citep{vulovic2013image}, which makes the particles almost invisible in Figure \ref{fig:real}. Our goal is to apply the proposed method to improve the damped contrast in the cryo-em image and reveal the particles from the background noise. Figure \ref{fig:real} visualizes the denoised images after applying the proposed algorithms trained for 100 epochs compared with BM3D. 
Our proposed algorithm can effectively reduce the background noise and reveal particles of interest, outperforming BM3D. Since some signal-dependent noises, e.g., the black ribbons, we apply a classic transformation technique, Variance Stabilizing Transform (VST), during our training process as suggested in \citet{makitalo2012optimal,zheng2020unsupervised}. After applying VST, such signal-dependent noises are further reduced.

To study the generalizability of the proposed algorithm, we further consider out-sample denoising, i.e., apply the trained denoiser to images outside the training set. We specifically apply the tuned models to other cryo-em images to assess the model out-sample performance. As BM3D is a non-learning-based method that cannot be evaluated on a external image, the out-sample performance is omitted in Figure \ref{fig:real}. From Figure \ref{fig:real}, we can see our  our method enhances the contrast of particles and reveals more information from the noisy image. Moreover, our trained model only takes $\sim 300$ seconds for patch-wise denoising on an external micrograph, whereas the BM3D requires $\sim 1500$ seconds per micrograph.

\paragraph{Synthetic Gaussian noise}

In this section, we evaluate the in-sample performance of the proposed procedure on one selected image in the training dataset $\textit{SET12}$ \citep{zhang2017beyond}. The chosen image is corrupted by additive Gaussian noise with various noise standard deviation $\sigma=20,25,30$. We compare our proposed method with two traditional denoising methods, Low-pass filter and Non-local mean (NLM) \citep{buades2005non}, and three learning-based methods, denoising CNN (DnCNN) \citep{zhang2017beyond}, deep image prior (DIP) \citep{ulyanov2018deep}, and NN+BM3D \citep{zheng2020unsupervised}; Specifically, DnCNN is a supervised method trained that requires clean/noisy image pairs; Low-pass and NLM do not provide out-sample denoiser; DIP, NN+BM3D, and our method only require single noisy images and yield out-sample denoisers. As suggested by \citet{quan2020self2self}, we choose a non-blind version of DIP guided by the estimated noise level for comparison. For a fair comparison, DnCNN is fine-tuned on the pre-trained model given the
provided single image and the NN+BM3D method is trained with the same neural network architecture as depicted in Figure \ref{fig:u-net}. All self-supervised learning methods are retrained on the selected single image for $K=100$ epochs after initialization.
\begin{table*}[!t]
    \caption{In-sample (top) and out-sample (bottom) performance in terms of PSNR/SSIM for synthetic additive Gaussian noise removal under different noise level }
    \vspace{0.15cm}
\centering 
\resizebox{.95\textwidth}{!}{
\begin{tabular}{cc|cc|c|ccc}
    \toprule
    \multicolumn{1}{l}{\textbf{In-sample}}&& \multicolumn{2}{c|}{Traditional methods} & \multicolumn{1}{c|}{Supervised method} 
    & \multicolumn{3}{c}{Self-supervised methods} \\
    $\sigma$ & {Noisy} & {Low-pass} & {NLM}  &{DnCNN} &{DIP}  & \multicolumn{1}{l}{NN+BM3D} & \multicolumn{1}{l}{NN+BM3D+T}  \\
    \midrule
    20 & 22.14dB/0.5212 & 29.89dB/0.9023 & 30.62dB/0.9041 & 29.98dB/0.8765 & 30.00dB/0.8823       & \textbf{30.22dB/0.9013} & {30.07dB/0.8932}  \\
     25 & 20.19dB/0.4255& 28.46dB/0.6238  & 29.28dB/0.8637   & 29.55dB/0.8729  & 28.62dB/0.8856      & 29.81dB/0.88611 & \textbf{30.44dB/0.9092}  \\
    30 & 18.72dB/0.3512 & 28.50dB/0.8623 & 28.06dB/0.8302   &   27.68dB/0.7578 & 28.23dB/0.8431  & 28.54dB/0.8461 & \textbf{29.34dB/0.8823}\\    
    \midrule
    \midrule
     \multicolumn{1}{l}{\textbf{Out-sample}}& $\sigma = 20$& {Low-pass} & {NLM}  &{DnCNN} &{DIP}  & \multicolumn{1}{l}{NN+BM3D} & \multicolumn{1}{l}{NN+BM3D+T}\\
      \midrule
     SET12&22.14dB/0.5921& NA & NA &\textbf{28.57dB/0.8832}& NA &27.01dB/0.8745&{27.16dB/0.8689}\\
     BSD68&22.13dB/0.6014& NA & NA &\textbf{28.00dB/0.8577}& NA &24.32dB/0.8424&{26.76dB/0.8434}\\
     \midrule 
     
     \multicolumn{1}{l}{\textbf{Out-sample}}& $\sigma = 25$& {Low-pass} & {NLM}  &{DnCNN} &{DIP}  & \multicolumn{1}{l}{NN+BM3D} & \multicolumn{1}{l}{NN+BM3D+T}\\
     \midrule
     SET12&20.19dB/0.5102& NA & NA &28.04dB/0.8721& NA &27.58dB/0.8732&\textbf{28.13dB/0.8921}\\
     BSD68&20.19dB/0.4834& NA & NA &\textbf{27.39dB/0.8478}& NA &26.43dB/0.8421&27.06dB/0.8456\\
      \midrule 
     \multicolumn{1}{l}{\textbf{Out-sample}}& $\sigma = 30$& {Low-pass} & {NLM}  &{DnCNN} &{DIP}  & \multicolumn{1}{l}{NN+BM3D} & \multicolumn{1}{l}{NN+BM3D+T}\\
     \midrule
     SET12&18.62dB/0.4211& NA & NA &26.44dB/0.8205& NA &26.67dB/0.8432&\textbf{27.08dB/0.8647}\\
     BSD68&18.61dB/0.4289& NA & NA &26.00dB/0.8017& NA &25.9049dB/0.8013&\textbf{26.41dB/0.8332}\\
    \bottomrule
    \end{tabular}}
    \label{tab:comparison}
\end{table*}

\begin{figure*}[!t]
    \centering
    
     \subfloat[\centering Truth (PSNR/SSIM)]{\includegraphics[width=.12\linewidth]{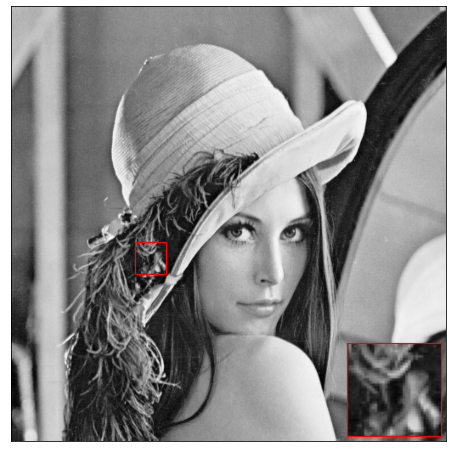}%
     }
     \subfloat[\centering Noisy (20.19dB/0.4255)]{\includegraphics[width=.12\linewidth]{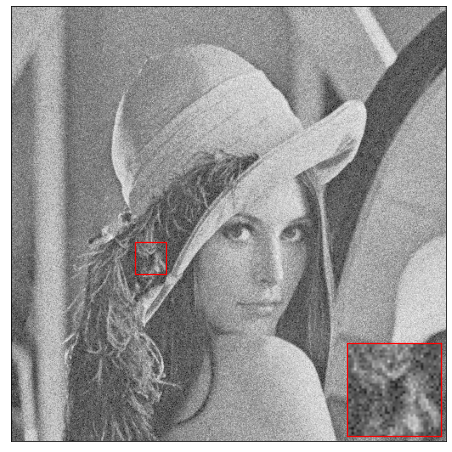}%
     }%
      \subfloat[\centering Low-pass (28.46dB/0.6238)]{\includegraphics[width=.12\linewidth]{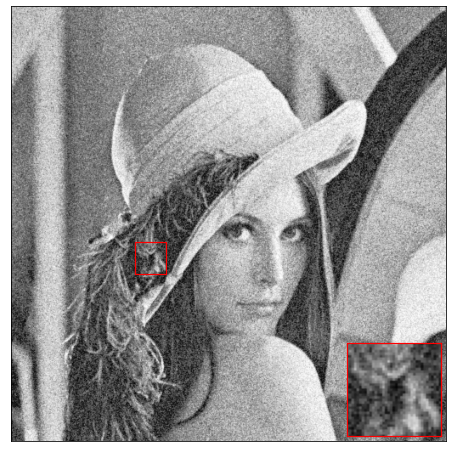}%
     }
         \subfloat[\centering NLM (29.28dB/0.8637)]{\includegraphics[width=.12\linewidth]{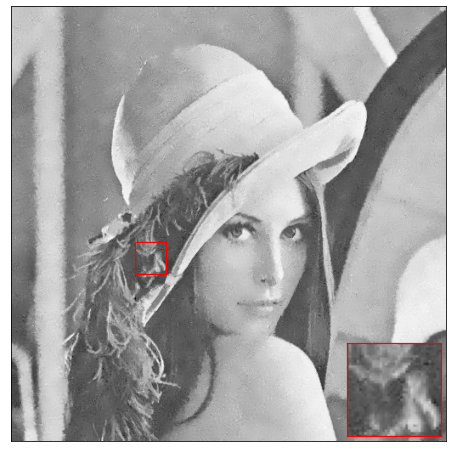}%
     }
     \subfloat[\centering DIP (28.62dB/0.8856)]{\includegraphics[width=.12\linewidth]{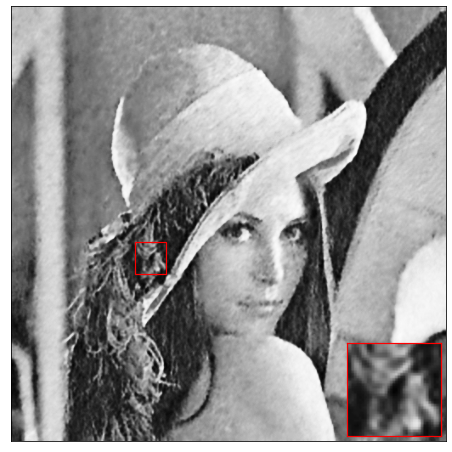}%
    }
      \subfloat[\centering DnCNN (29.55dB/0.8729)]{\includegraphics[width=.12\linewidth]{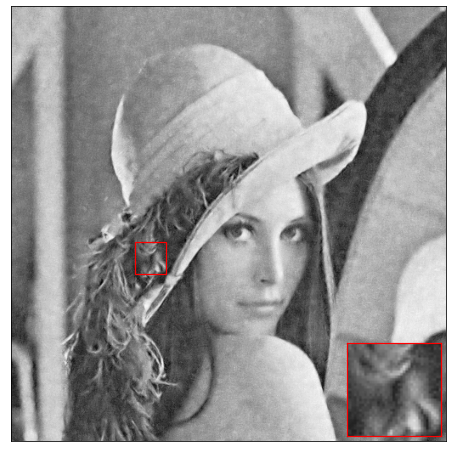}%
    }
    \subfloat[\centering NN+BM3D (29.81dB/0.8861)]{\includegraphics[width=.12\linewidth]{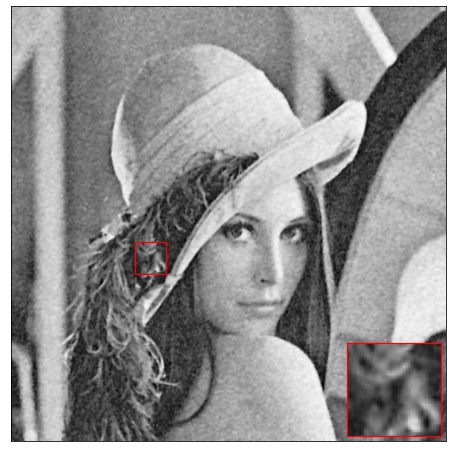}%
    }
    \subfloat[\centering NN+BM3D+T (30.44dB/0.9092)]{\includegraphics[width=.12\linewidth]{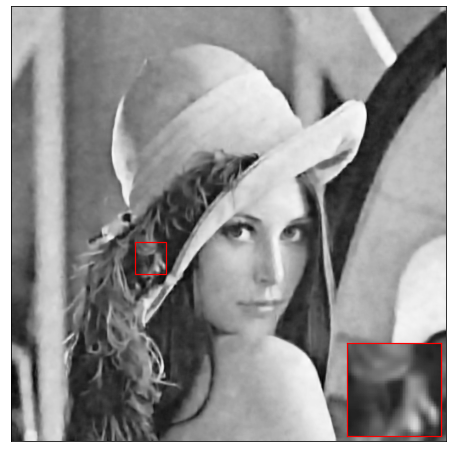}%
    }\\
     \subfloat[\centering Truth (PSNR/SSIM)]{\includegraphics[width=.12\linewidth]{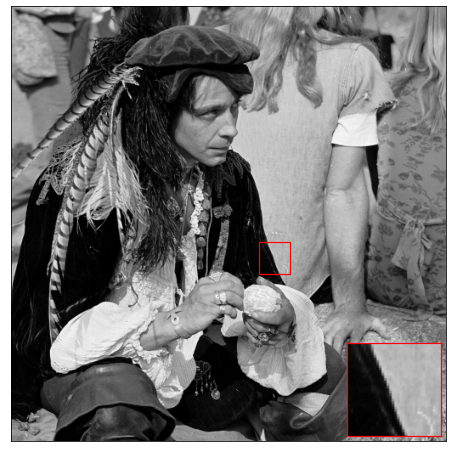}%
     }
     \subfloat[\centering Noisy (20.18dB/0.4701)]{\includegraphics[width=.12\linewidth]{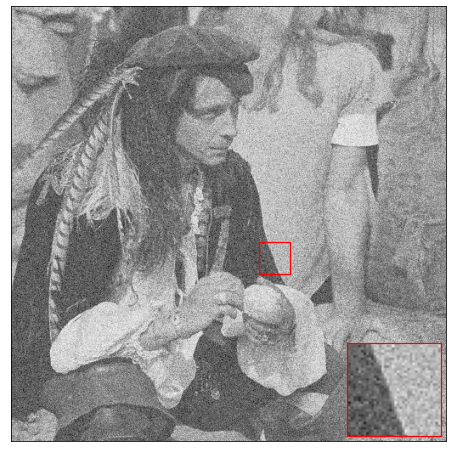}%
     }%
      \subfloat[\centering Low-pass (27.69dB/0.8534)]{\includegraphics[width=.12\linewidth]{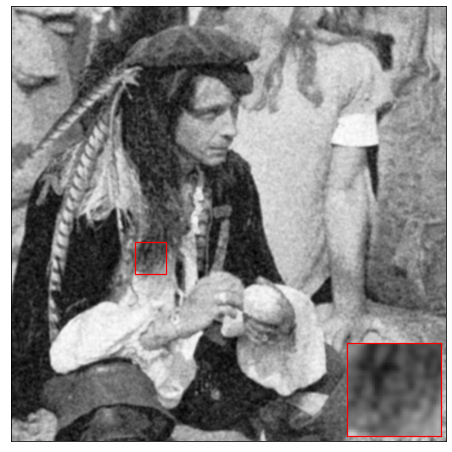}%
     }
         \subfloat[\centering NLM (28.03dB/0.8402)]{\includegraphics[width=.12\linewidth]{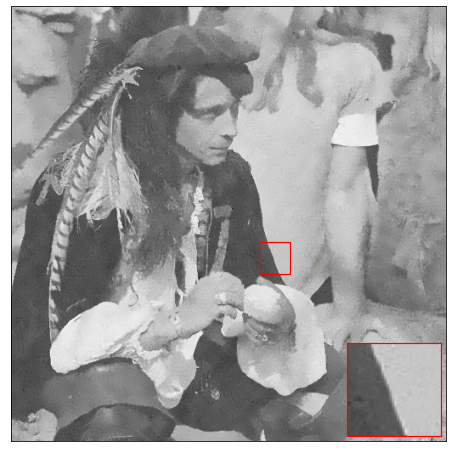}%
     }
     \subfloat[\centering DIP (27.53dB/0.8208)]{\includegraphics[width=.118\linewidth]{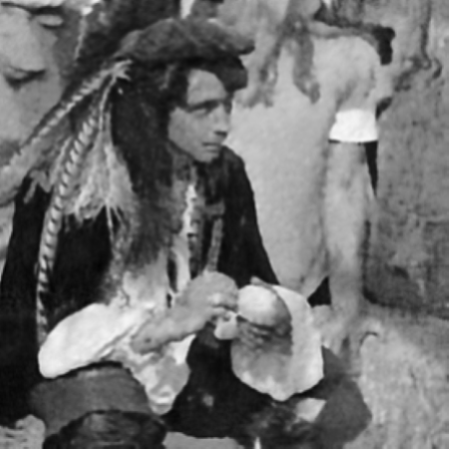}%
    }
      \subfloat[\centering DnCNN (27.61dB/0.8392)]{\includegraphics[width=.12\linewidth]{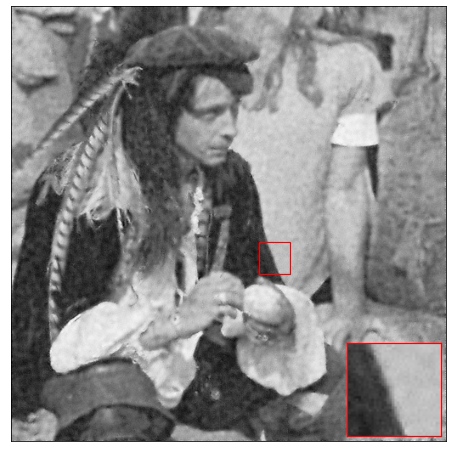}%
    }
    \subfloat[\centering NN+BM3D (28.47dB/0.8652)]{\includegraphics[width=.12\linewidth]{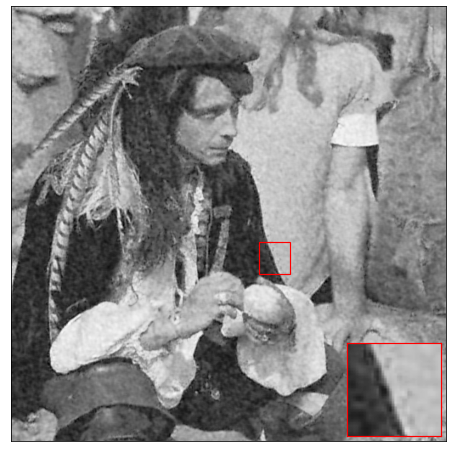}%
    }
    \subfloat[\centering NN+BM3D+T (28.67dB/0.8701)]{\includegraphics[width=.12\linewidth]{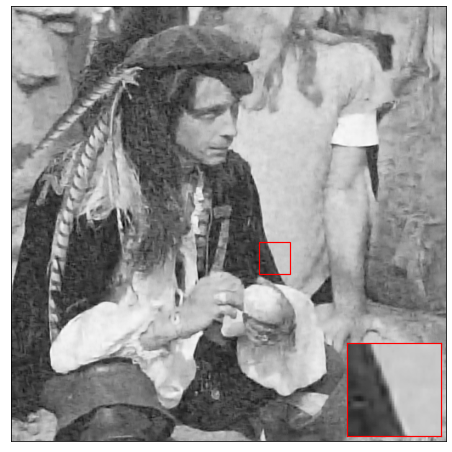}%
    }
    \caption{Visual comparisons of our method against other competing methods in terms of in-sample performance from dataset $\textit{SET12}$ coupled with PSNR and SSIM. See more in-sample comparisons in the supplementary materials.}
    \label{fig:lena}
\end{figure*}

The results are assessed in both peak signal-to-noise ratio (PSNR) and structured similarity index (SSIM) computed by the built-in Python functions in the \texttt{skimage} package. Figure \ref{fig:lena} illustrates one example of the in-sample performance in the dataset $\textit{SET12}$ when $\sigma=25$; see Table \ref{tab:comparison} for a comprehensive in-sample comparison under different noise levels. We can see (i) learning-based methods perform better and generalize better to out-samples; (ii) our proposed method outperforms its precedents NN+BM3D in terms of PSNR/SSIM under moderate noise levels. However, when the noise is diminishing in intensity, i.e., $\sigma = 20$, our method is subject to a minor quality drop -- this is reasonable as the image presents less noise, our low-rank approximation might filter out some useful information.

\paragraph{Synthetic Poisson-Gaussian noise}

\begin{figure}[!t]
    \centering
    \subfloat[\scriptsize 27.22dB/0.7552]{\includegraphics[width=.23\linewidth]{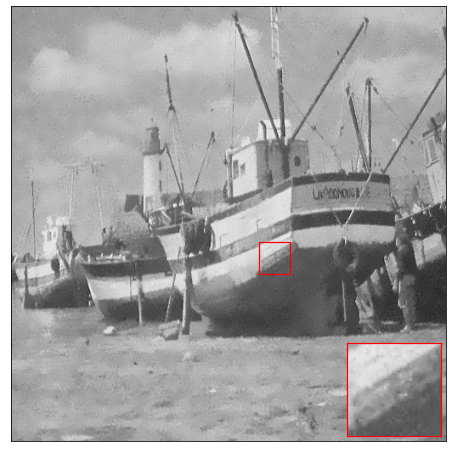}%
    }
    \subfloat[\scriptsize 31.22dB/0.9182]{\includegraphics[width=.23\linewidth]{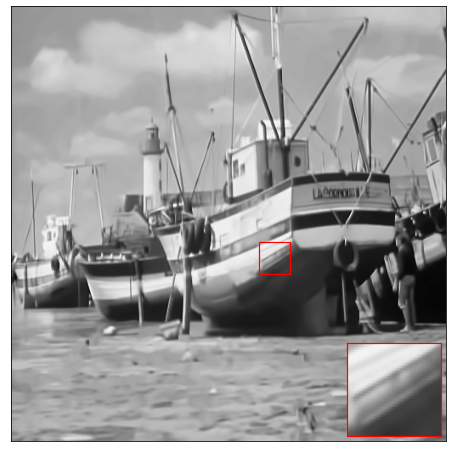}%
    }
    \subfloat[\scriptsize 29.81dB/0.9013]{\includegraphics[width=.23\linewidth]{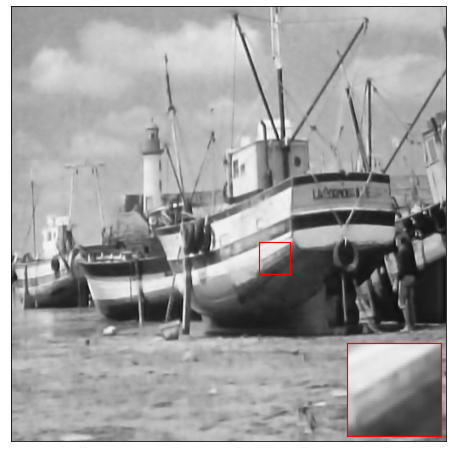}%
    }
     \subfloat[\scriptsize 31.77dB/0.9322]{\includegraphics[width=.23\linewidth]{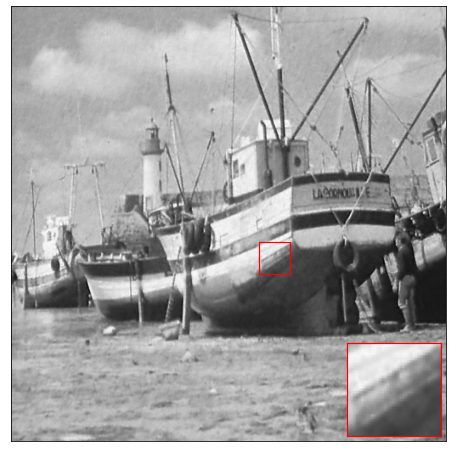}%
    }
    \caption{A visual comparison of (a) Poisson-Gaussian-noisy image, (b) denoised by BM3D, (c) denoised by NN+BM3D+T, and (d) denoised by NN+BM3D+VST+T ($\eta = 0.5, \rho = 100$ and $S_D = 200$).}
    \label{fig:poisson}
\end{figure}

We also test the performance of the proposed method on images corrupted by Poisson-Gaussian distributed noise \citep{foi2008practical,wang2020practical}. First, we synthesize the noisy images by generating Poisson-Gaussian pixel values based on one selected single image. Next, we apply the proposed method coupled with the VST technique on the synthetic noisy image for comparison. The visual and quantitative evaluation in Figure \ref{fig:poisson} shows our method works effectively for Poisson-Gaussian noise removal.

\paragraph{Out-sample performance}

\begin{figure}[!t]
    \centering
    \subfloat[]{\includegraphics[width=.48\linewidth]{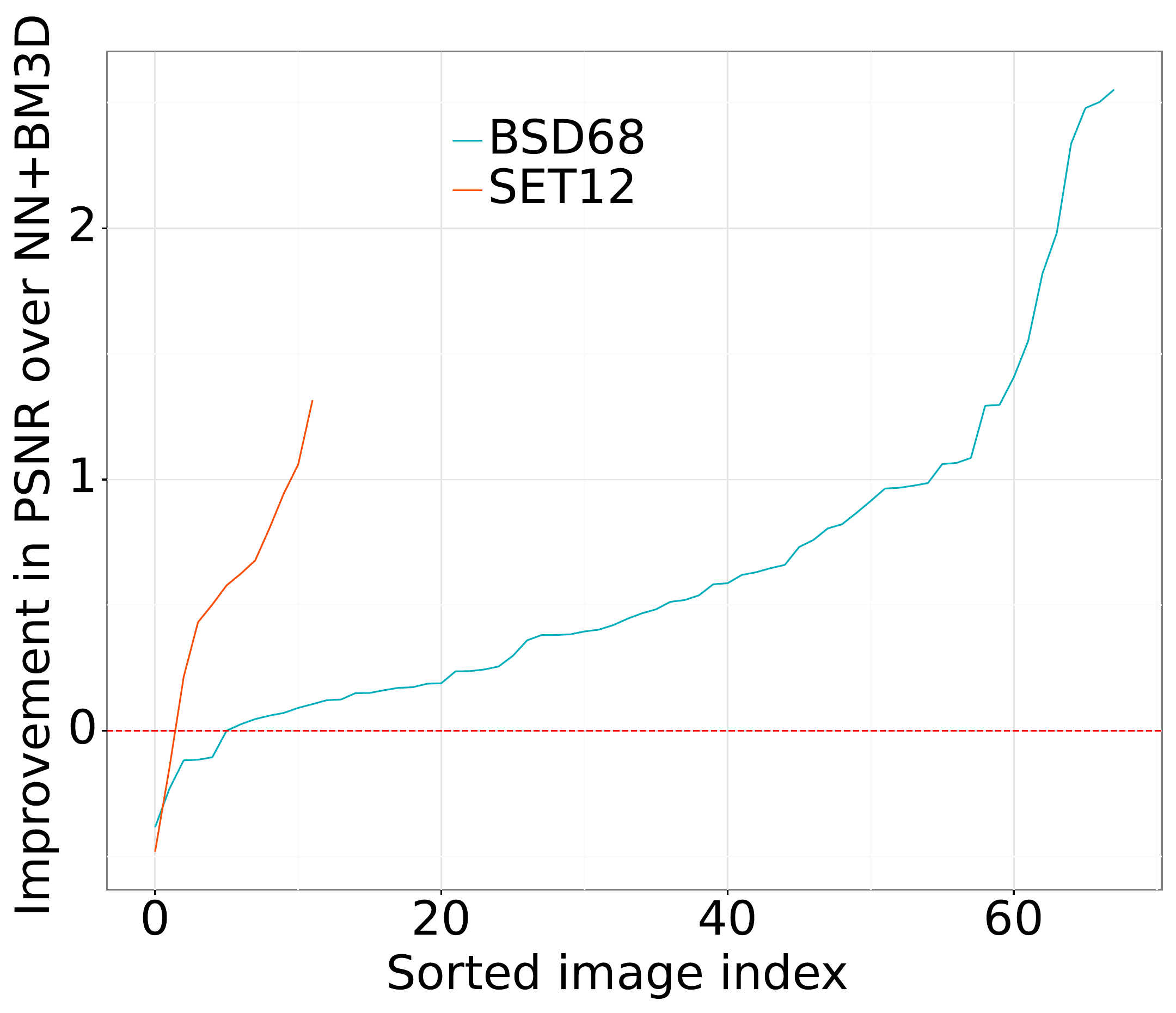}%
    \label{fig:compare1}
    }
    \subfloat[]{\includegraphics[width=.48\linewidth]{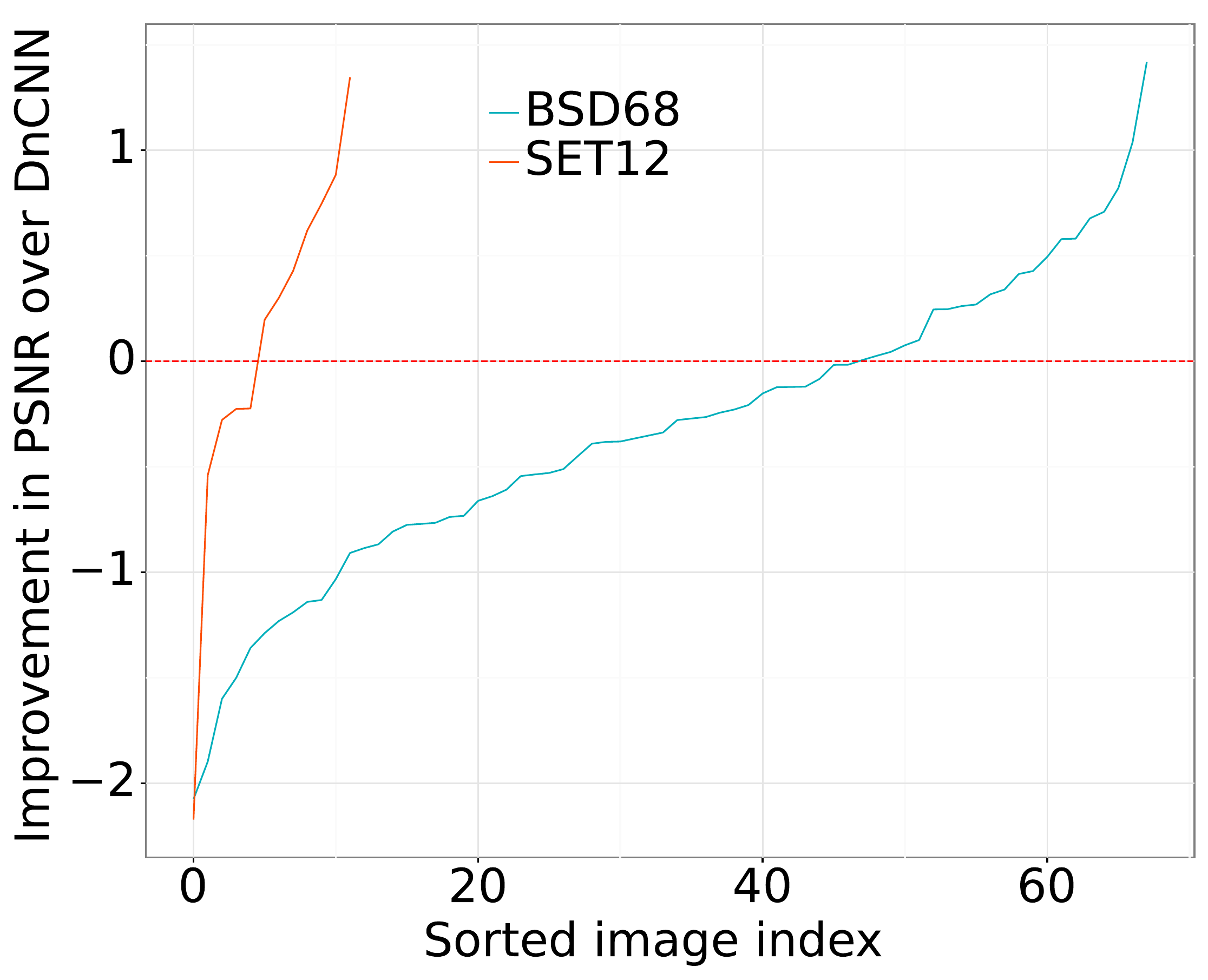}%
    \label{fig:compare2}
    }
    \caption{Out-sample performance profile of our algorithm on two datasets (\textit{SET12} and \textit{BSD68}) when $\sigma=25$ compared to (a) the single-image-based denoiser NN+BM3D and (b) the supervised learning-based denoiser DnCNN. Note that the our denoiser is trained on \textit{LENA} but tested on the other images.}
    \label{fig:generalization}
\end{figure}

We further compare the out-sample performance of our method to the one without weight distortions (i.e., ablation analysis) (Figure \ref{fig:compare1}) and a supervised learning-based denoiser DnCNN (Figure \ref{fig:compare2}) when $\sigma=25$; see Table \ref{tab:comparison} for a comprehensive out-sample comparison. Specifically, we train the denoisers based on the single noisy image \textit{LENA} and evaluate the out-sample performance on the remaining test images in datasets \textit{SET12} and \textit{BSD68} \citep{MartinFTM01} corrupted by additive Gaussian noises under the same noise level as the training image.

In addition, we present the out-sample performance under misaligned noise levels in Table \ref{tab:out_misaligned}, where the performance is evaluated on the test images (e.g., BSD68) with different noise levels from the training image (e.g., one noisy image from SET12). This demonstrates the advantages of weight distortion to the model generalization ability.

\begin{table}[!t]
    \caption{Out-sample performance in terms of PSNR/SSIM for synthetic additive Gaussian noise removal under misaligned noise level
    \label{tab:out_misaligned}}
    \vspace{0.15cm}
\centering 
%
    \begin{tabular}{lllll}
    \toprule
    \multicolumn{1}{l}{Training} &  \multicolumn{1}{l}{Test}  & \multicolumn{1}{l}{DnCNN} & \multicolumn{1}{l}{NN+BM3D} & \multicolumn{1}{l}{NN+BM3D+T} \\
    \midrule
    {$\sigma=20$} & $\sigma=20$ &  \textbf{28.00dB/0.8621}    &24.32dB/0.8433      &  26.76dB/0.8421\\
          & $\sigma=25$ &  \textbf{26.54dB/0.8145}    &      23.97dB/0.8123 &  26.07dB/0.8126 \\
          & $\sigma=30$ &  24.36dB/0.7234     &  23.57dB/0.7821     & \textbf{25.37dB/0.7848} \\
    \midrule
    $\sigma=25$ & $\sigma=20$ &      \textbf{28.14dB/0.8745} &   27.06dB/0.8579    & 27.69dB/0.8723\\
          & $\sigma=25$ & \textbf{27.39dB/0.8548}      &  26.34dB/0.8421     & 27.06dB/0.8532  \\
          & $\sigma=30$ &  26.08dB/0.8011   &  25.54dB/0.8046   & \textbf{26.18dB/0.8089}\\
    \midrule
    $\sigma=30$ & $\sigma=20$ &  \textbf{28.00dB/0.8631}     &  27.84dB/0.8746    & 27.61dB/0.8639 \\
          & $\sigma=25$ &   26.54dB/0.8109    &    26.92dB/0.8411   & \textbf{27.07dB/0.8541} \\
          & $\sigma=30$ &   26.00dB/0.8002    &  25.90dB/0.8033     &  \textbf{26.41dB/0.8329} \\
    \bottomrule
    \end{tabular}%
\end{table}

The results demonstrate that our proposed network can achieve better results when being evaluated with an external-image tuned parameters on average, is even competitive with the supervised denoiser (e.g., DnCNN) across all considered (aligned or misaligned) noise levels. Such results manifest the benefit of adding a Tucker low-rank approximation for the weight tensor in terms of improving generalization despite a minor quality drops of in-sample performance. See more out-sample experiments in the supplementary material.

\subsection{Discussions}\label{sec:discuss}
\paragraph{Advantages of low-rank approximation}

\begin{figure}[!t]
\centering

\subfloat[]{\includegraphics[width=15em]{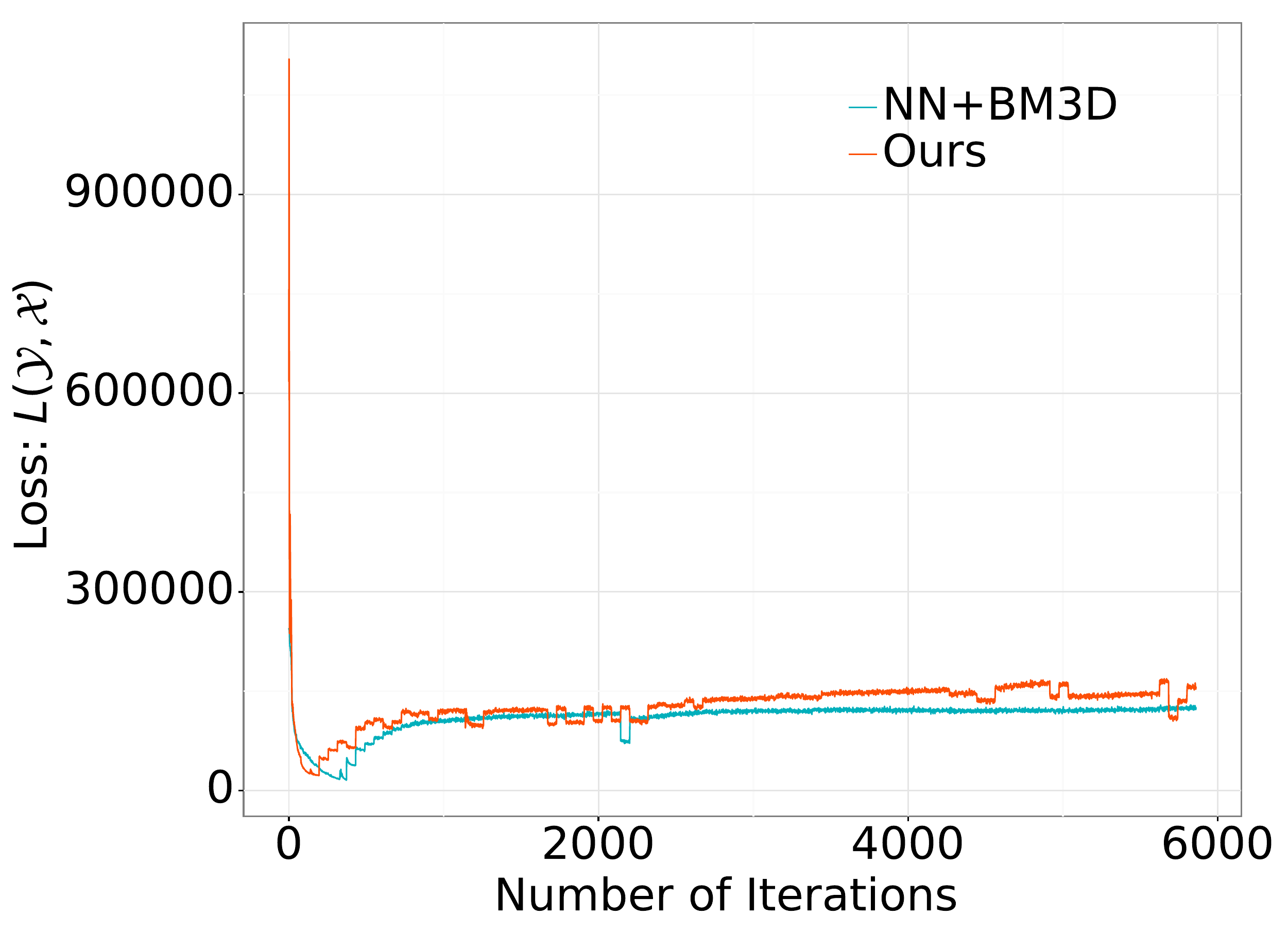}%
    \label{fig:loss:(a)}
    }
\subfloat[]{\includegraphics[width=15em]{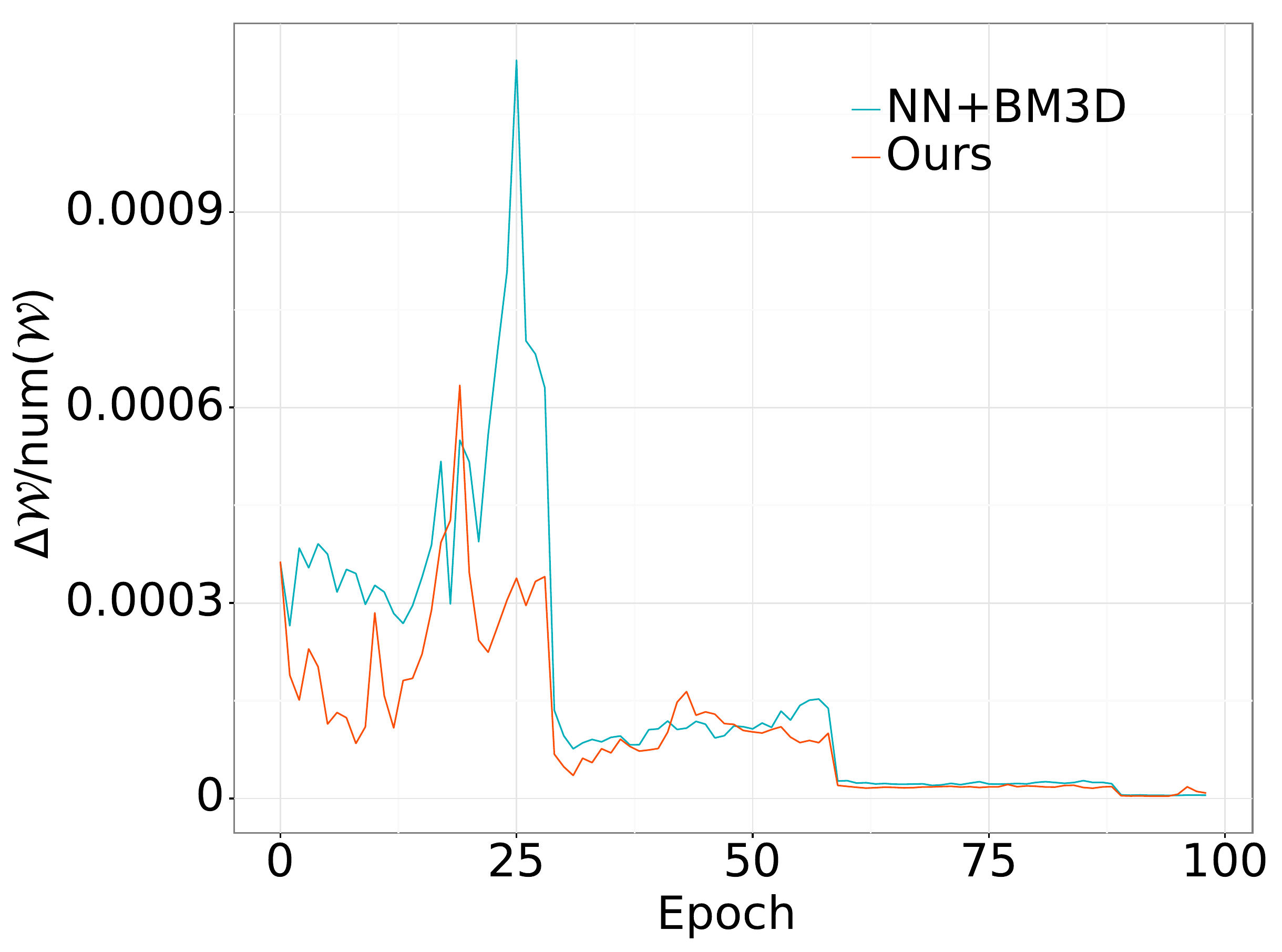}%
    \label{fig:weight:(a)}
    }\\
    \subfloat[]{\includegraphics[width=15em]{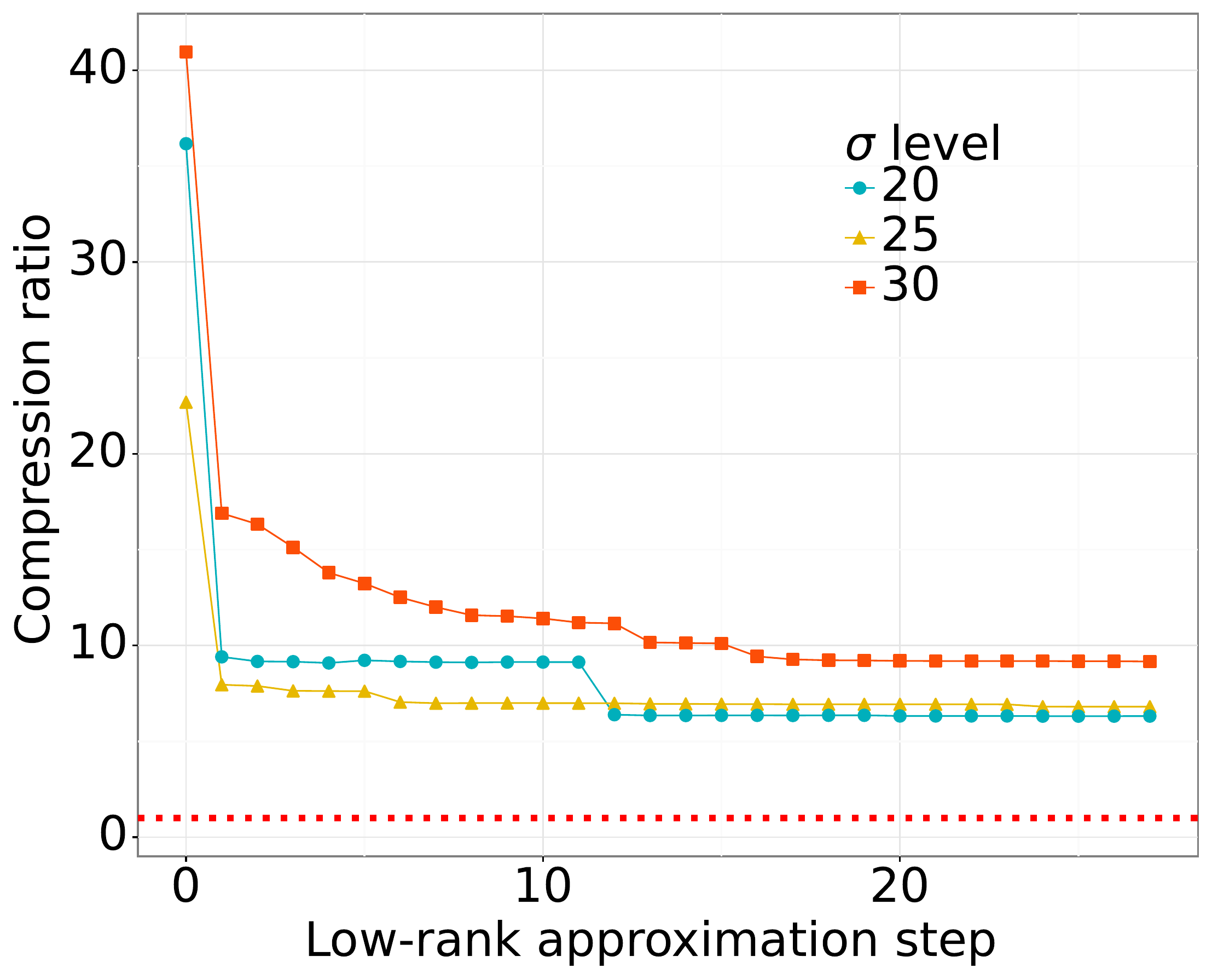}%
    \label{fig:cr:(a)}
    }
    \subfloat[]{\includegraphics[width=15em]{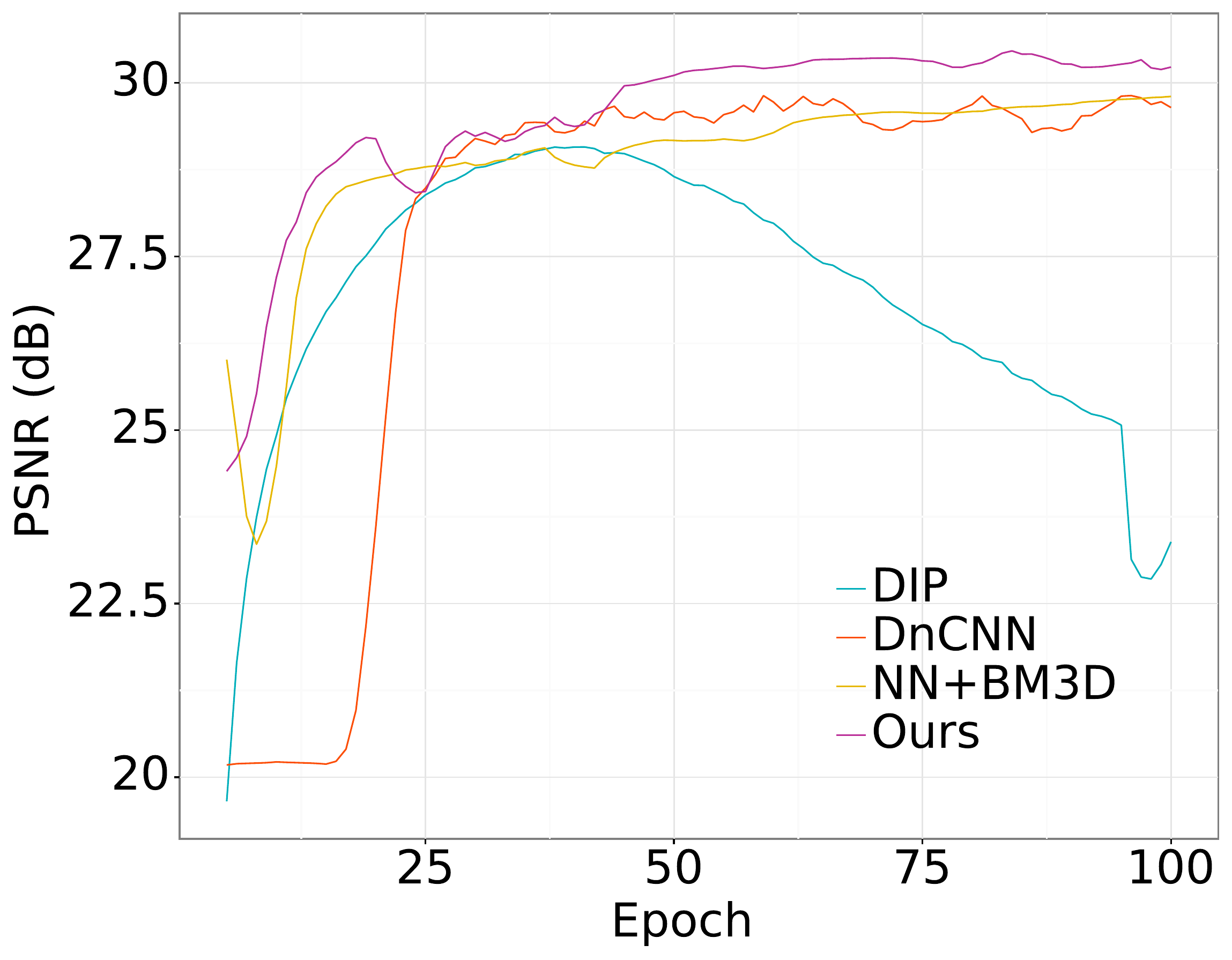}%
    \label{fig:g:(a)}
    }
    \caption{(a) Training loss $l\{f_{\theta^{(k),i}}(\mathcal{Y}),\mathcal{Y}\}$ of NN+BM3D and ours versus number of iterations when $\sigma =25$; (b) Average changes in weight tensors $\Delta W = \|\mathcal{W}^{k} - \mathcal{W}^{k-1}\|_F$ of NN+BM3D and ours versus epochs when $\sigma =25$; (c) Average compression ratio of parameters at each distortion step in NN+BM3D+T when $\sigma = 20, 25, 30$; (d) Average PSNR of $5$ initial starts for training single image $\textit{LENA}$ over epochs.}
    \label{fig:loss_weight}
\end{figure}

To assess the benefits of low-rank tensor approximation on convergence, we present in Figure \ref{fig:loss_weight} the loss function $l\{f_{\theta^{(k),i}}(\mathcal{Y}), \mathcal{Y}\}$ over each iteration $i$ within epoch $k$, average changes of the weight kernels in Frobenius norm versus epochs (average difference of weight tensor is measured by $\Delta \mathcal{W}/(d^2 s\widehat{s})$ for $\mathcal{W}\in \mathbb{R}^{d\times d\times s \times \widehat{s}}$), and the average compression ratio (over layers) defined in \eqref{eq:CR} at each low-rank approximation step for all $\sigma$ levels. One can observe that (i) the loss functions in both models decrease significantly after a few iterations; (ii) changes of weights tend to fluctuate less when coupled with the weight distortion (Figure \ref{fig:weight:(a)}), and the embedded low-rank Tucker approximation does not noticeably affect the final loss value (Figure \ref{fig:loss:(a)}); (iii) average compression ratios are significantly reduced after one low-rank approximation step but remain above $1$ henceforth under all noise levels (Figure \ref{fig:cr:(a)}). Supported by the empirical evidence, it can be concluded that applying the low-rank approximation to the weight tensors can be beneficial to streamline the neural network structure, leading to a flatter local minima without notable loss increment.

We further consider the computational complexity of training. Our proposed method takes $\sim 330$ seconds per epoch to denoise one image of size $256\times256\times 1$, which requires less variant training time as implied by its smaller standard deviation. See Table \ref{tab:time}.

\begin{table}[!t]
    \centering
     \caption{Average training time in seconds per epoch (standard deviation in parentheses) and inference time in seconds per image for learning based denoisers
    }
    \vspace{0.25cm}
    \begin{tabular}{cccccc}
    \hline
        Methods & DnCNN & DIP & NN+BM3D & NN+BM3D+T\\
        \hline  
        Training time&
          283.83 (78.1) & 495.86 (6.8) & 340.23 (9.0) & \textbf{332.71} (\textbf{1.2})\\
         Inference time & 4.99 & 2.52 & 2.28 & 2.32\\
         \hline
    \end{tabular}
    \label{tab:time}
\end{table}

\paragraph{Sensitivity analysis}

We also assess the robustness of model performance under different initialization weight values ($\boldsymbol{\theta}^0$) and hyper-parameters ($\rho$, $\eta$ and $S_D$). As the gradient-based optimization is known to be sensitive to initialization \citep{bubeck2014convex}, we first investigate the model robustness under different weight initializations based on in-sample PSNR stabilizability. Specifically, we re-run the learning-based methods (DIP, DnCNN, NN+BM3D, and ours) for $5$ times after $100$ epochs on one single image, \textit{LENA}, with the Xavier weight initialization strategy in \citet{glorot2010understanding}. Figure \ref{fig:g:(a)} displays the average training PSNR values against the epochs, which shows our algorithm reaches stability with the highest average PSNR after $100$ epochs and shows its robustness to the randomness of the initialization.

\begin{figure}[!t]
    \centering
    \includegraphics[width=.3\linewidth]{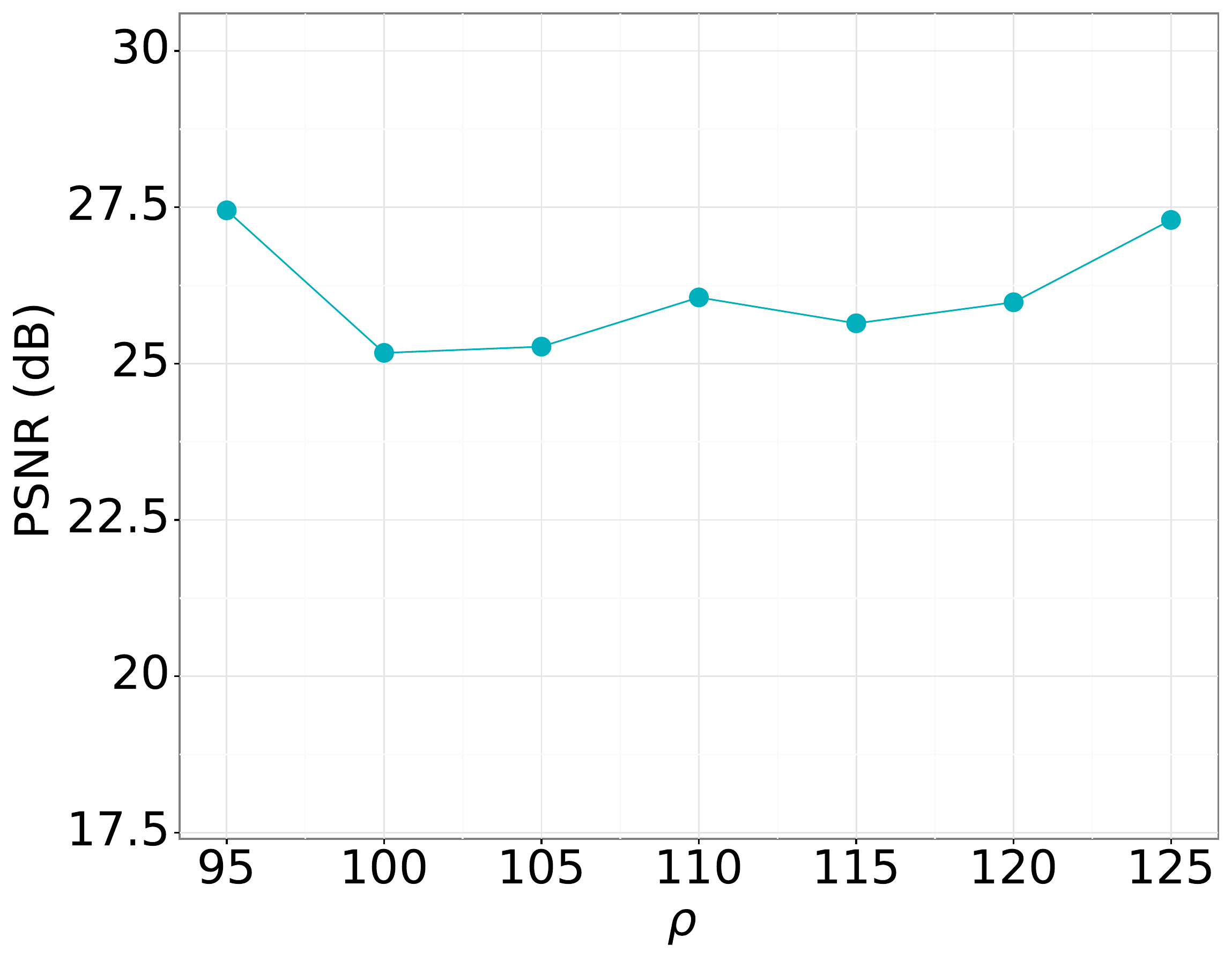}
    \includegraphics[width=.3\linewidth]{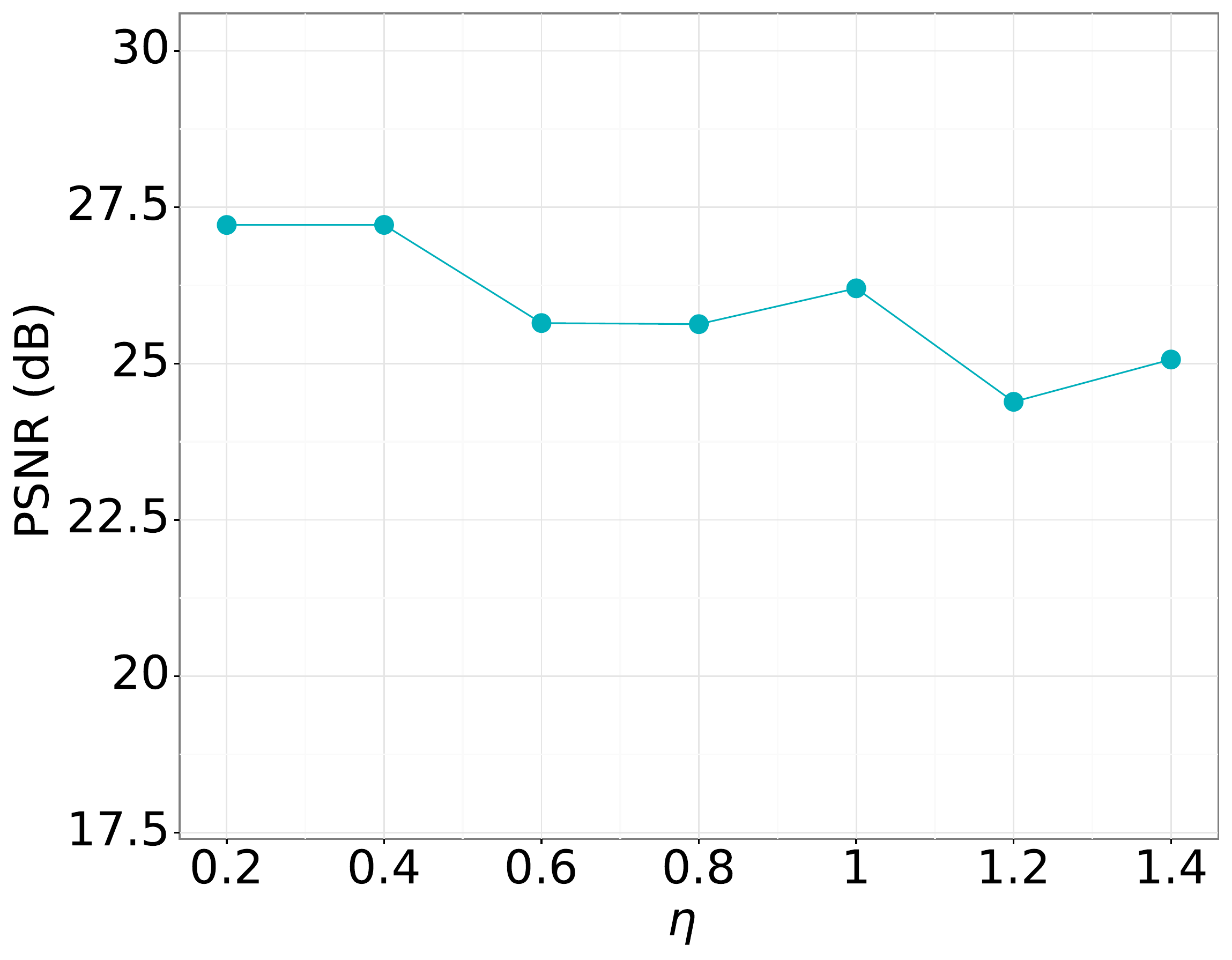}
    \includegraphics[width=.3\linewidth]{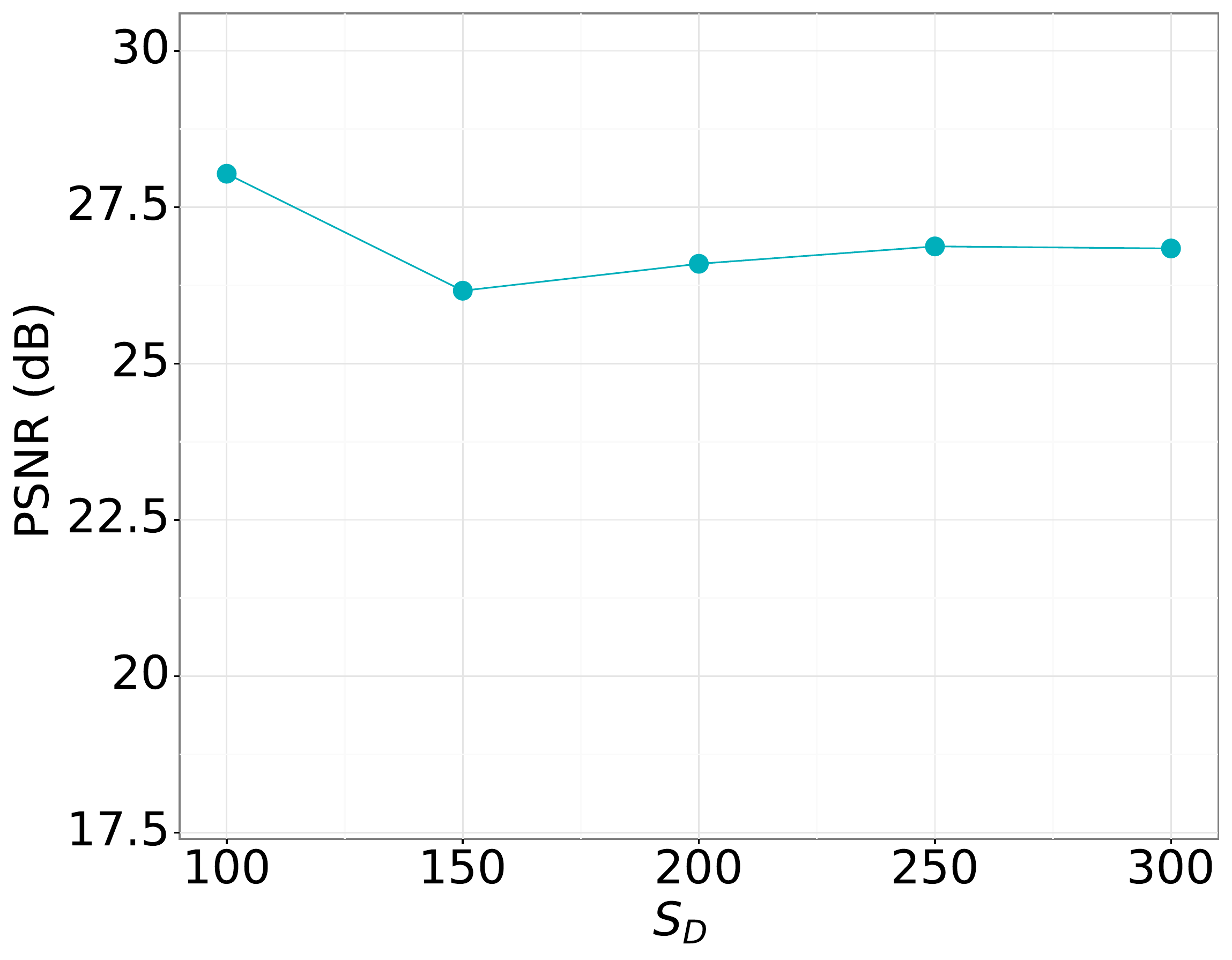}
    \caption{Sensitivity of hyper-parameters $\rho$, $\eta$ and $S_D$.}
    \label{fig:sen}
\end{figure}

Next, we perform sensitivity analysis on hyper-parameters for our proposed algorithm. We calculate the average PSNR on the dataset $\textit{SET12}$ by fixing $\eta=0.5, S_D = 200$ and varying $\rho$ from $95$ to $125$ (Figure \ref{fig:sen}, left), by fixing $\rho=100, S_D=100$ and varying $\eta$ from $0.2$ to $1.4$ (Figure \ref{fig:sen}, middle), and by fixing $\rho=100,\eta=0.5$ and varying $S_D$ from $100$ to $300$ (Figure \ref{fig:sen}, right). The resulting PSNR curves are stable in all scenarios, which shows our model is robust to different choices of hyper-parameters.

\section{Conclusion}\label{sec:discussion}
    In this paper, we propose to substitute the weight kernels in the CNN-based network with their approximated low-rank tensors in the original learning structure. By combining this technique with the state-of-the-art single-image denoising methods, the resulting network is equipped with more flexibility and improved generalization ability. In addition, with the aid of VBMF for optimal rank selection, our algorithm is data-driven and automatic as it requires little manual interference and achieves the end-to-end fashion. ow-rank weight approximation for multi-channel images (e.g., RGB-colored images) might require more carefulness in rank selection to deal with the inter-channel correlated noises, where the column-wise independence required by the VBMF algorithm is not satisfied. Empirical evidence based on the synthetic noisy images and real-world noisy images demonstrates the advantages of proposed weight distortion compared with various learning-based and non-learning-based methods. As the column-wise independence required by the VBMF algorithm is not satisfied for multi-channel images (e.g., RGB-colored images), low-rank weight approximation for multi-channel images might require more careful handling of inter-channel correlated noises. An extension to handling rank selections under inter-channel correlated noises is one of our future directions (e.g., incorporating the estimation of noise covariance across channels \citep{dong2018color}). Further tasks in image processing, such as deblurring, segmentation, and classification, are to be explored in future research as well.

\section*{Acknowledgement}
The authors thank Priyamvada Acharya for providing the cryo-EM dataset of SARS-CoV-2 2P protein and Pixu Shi for helpful discussions.
\bibliographystyle{apalike}
\bibliography{main}

\begin{thebibliography}{}

\bibitem[Aharon et~al., 2006]{aharon2006k}
Aharon, M., Elad, M., and Bruckstein, A. (2006).
\newblock K-svd: An algorithm for designing overcomplete dictionaries for
  sparse representation.
\newblock {\em IEEE Transactions on signal processing}, 54(11):4311--4322.

\bibitem[Arora et~al., 2018]{arora2018stronger}
Arora, S., Ge, R., Neyshabur, B., and Zhang, Y. (2018).
\newblock Stronger generalization bounds for deep nets via a compression
  approach.
\newblock In {\em International Conference on Machine Learning}, pages
  254--263. PMLR.

\bibitem[Baldassi et~al., 2020]{baldassi2020shaping}
Baldassi, C., Pittorino, F., and Zecchina, R. (2020).
\newblock Shaping the learning landscape in neural networks around wide flat
  minima.
\newblock {\em Proceedings of the National Academy of Sciences},
  117(1):161--170.

\bibitem[Bepler et~al., 2020]{bepler2020topaz}
Bepler, T., Kelley, K., Noble, A.~J., and Berger, B. (2020).
\newblock Topaz-denoise: general deep denoising models for cryoem and cryoet.
\newblock {\em Nature communications}, 11(1):1--12.

\bibitem[Boyd et~al., 2011]{boyd2011distributed}
Boyd, S., Parikh, N., and Chu, E. (2011).
\newblock {\em Distributed optimization and statistical learning via the
  alternating direction method of multipliers}.
\newblock Now Publishers Inc.

\bibitem[Buades et~al., 2005a]{buades2005non}
Buades, A., Coll, B., and Morel, J.-M. (2005a).
\newblock A non-local algorithm for image denoising.
\newblock In {\em 2005 IEEE Computer Society Conference on Computer Vision and
  Pattern Recognition (CVPR'05)}, volume~2, pages 60--65. IEEE.

\bibitem[Buades et~al., 2005b]{buades2005review}
Buades, A., Coll, B., and Morel, J.-M. (2005b).
\newblock A review of image denoising algorithms, with a new one.
\newblock {\em Multiscale modeling \& simulation}, 4(2):490--530.

\bibitem[Buban et~al., 2010]{buban2010high}
Buban, J.~P., Ramasse, Q., Gipson, B., Browning, N.~D., and Stahlberg, H.
  (2010).
\newblock High-resolution low-dose scanning transmission electron microscopy.
\newblock {\em Journal of electron microscopy}, 59(2):103--112.

\bibitem[Bubeck, 2014]{bubeck2014convex}
Bubeck, S. (2014).
\newblock Convex optimization: Algorithms and complexity.
\newblock {\em arXiv preprint arXiv:1405.4980}.

\bibitem[Bulat et~al., 2019]{bulat2019matrix}
Bulat, A., Kossaifi, J., Tzimiropoulos, G., and Pantic, M. (2019).
\newblock Matrix and tensor decompositions for training binary neural networks.
\newblock {\em arXiv preprint arXiv:1904.07852}.

\bibitem[Cha and Moon, 2019]{cha2019fully}
Cha, S. and Moon, T. (2019).
\newblock Fully convolutional pixel adaptive image denoiser.
\newblock In {\em Proceedings of the IEEE/CVF International Conference on
  Computer Vision}, pages 4160--4169.

\bibitem[Chen et~al., 2015a]{chen2015efficient}
Chen, G., Zhu, F., and Ann~Heng, P. (2015a).
\newblock An efficient statistical method for image noise level estimation.
\newblock In {\em Proceedings of the IEEE International Conference on Computer
  Vision}, pages 477--485.

\bibitem[Chen et~al., 2015b]{chen2015compressing}
Chen, W., Wilson, J., Tyree, S., Weinberger, K., and Chen, Y. (2015b).
\newblock Compressing neural networks with the hashing trick.
\newblock In {\em International conference on machine learning}, pages
  2285--2294. PMLR.

\bibitem[Dabov et~al., 2007]{dabov2007image}
Dabov, K., Foi, A., Katkovnik, V., and Egiazarian, K. (2007).
\newblock Image denoising by sparse 3-d transform-domain collaborative
  filtering.
\newblock {\em IEEE Transactions on image processing}, 16(8):2080--2095.

\bibitem[De~Lathauwer et~al., 2000]{de2000best}
De~Lathauwer, L., De~Moor, B., and Vandewalle, J. (2000).
\newblock On the best rank-1 and rank-(r 1, r 2,..., rn) approximation of
  higher-order tensors.
\newblock {\em SIAM Journal on Matrix Analysis and Applications},
  21(4):1324--1342.

\bibitem[Dong et~al., 2018]{dong2018color}
Dong, L., Zhou, J., and Dai, T. (2018).
\newblock Color image noise covariance estimation with cross-channel image
  noise modeling.
\newblock In {\em 2018 IEEE International Conference on Multimedia and Expo
  (ICME)}, pages 1--6. IEEE.

\bibitem[Fan et~al., 2019]{fan2019brief}
Fan, L., Zhang, F., Fan, H., and Zhang, C. (2019).
\newblock Brief review of image denoising techniques.
\newblock {\em Visual Computing for Industry, Biomedicine, and Art},
  2(1):1--12.

\bibitem[Foi et~al., 2008]{foi2008practical}
Foi, A., Trimeche, M., Katkovnik, V., and Egiazarian, K. (2008).
\newblock Practical poissonian-gaussian noise modeling and fitting for
  single-image raw-data.
\newblock {\em IEEE Transactions on Image Processing}, 17(10):1737--1754.

\bibitem[Glorot and Bengio, 2010]{glorot2010understanding}
Glorot, X. and Bengio, Y. (2010).
\newblock Understanding the difficulty of training deep feedforward neural
  networks.
\newblock In {\em Proceedings of the thirteenth international conference on
  artificial intelligence and statistics}, pages 249--256. JMLR Workshop and
  Conference Proceedings.

\bibitem[Gu et~al., 2014]{gu2014weighted}
Gu, S., Zhang, L., Zuo, W., and Feng, X. (2014).
\newblock Weighted nuclear norm minimization with application to image
  denoising.
\newblock In {\em Proceedings of the IEEE conference on computer vision and
  pattern recognition}, pages 2862--2869.

\bibitem[Han et~al., 2015a]{han2015deep}
Han, S., Mao, H., and Dally, W.~J. (2015a).
\newblock Deep compression: Compressing deep neural networks with pruning,
  trained quantization and huffman coding.
\newblock {\em arXiv preprint arXiv:1510.00149}.

\bibitem[Han et~al., 2015b]{han2015learning}
Han, S., Pool, J., Tran, J., and Dally, W.~J. (2015b).
\newblock Learning both weights and connections for efficient neural networks.
\newblock {\em arXiv preprint arXiv:1506.02626}.

\bibitem[Hinton et~al., 2015]{hinton2015distilling}
Hinton, G., Vinyals, O., and Dean, J. (2015).
\newblock Distilling the knowledge in a neural network.
\newblock {\em arXiv preprint arXiv:1503.02531}.

\bibitem[Huang et~al., 2017]{huang2017highly}
Huang, H., Ni, L., Wang, K., Wang, Y., and Yu, H. (2017).
\newblock A highly parallel and energy efficient three-dimensional multilayer
  cmos-rram accelerator for tensorized neural network.
\newblock {\em IEEE Transactions on Nanotechnology}, 17(4):645--656.

\bibitem[Huang et~al., 2021]{huang2021neighbor2neighbor}
Huang, T., Li, S., Jia, X., Lu, H., and Liu, J. (2021).
\newblock Neighbor2neighbor: Self-supervised denoising from single noisy
  images.
\newblock In {\em Proceedings of the IEEE/CVF Conference on Computer Vision and
  Pattern Recognition}, pages 14781--14790.

\bibitem[Hubara et~al., 2016]{hubara2016binarized}
Hubara, I., Courbariaux, M., Soudry, D., El-Yaniv, R., and Bengio, Y. (2016).
\newblock Binarized neural networks.
\newblock {\em Advances in neural information processing systems}, 29.

\bibitem[Kim et~al., 2016]{kim2015compression}
Kim, Y.-D., Park, E., Yoo, S., Choi, T., Yang, L., and Shin, D. (2016).
\newblock Compression of deep convolutional neural networks for fast and low
  power mobile applications.
\newblock {\em International Conference on Learning Representations}.

\bibitem[Kingma and Ba, 2014]{kingma2014adam}
Kingma, D.~P. and Ba, J. (2014).
\newblock Adam: A method for stochastic optimization.
\newblock {\em arXiv preprint arXiv:1412.6980}.

\bibitem[Kolda, 2001]{kolda2001orthogonal}
Kolda, T.~G. (2001).
\newblock Orthogonal tensor decompositions.
\newblock {\em SIAM Journal on Matrix Analysis and Applications},
  23(1):243--255.

\bibitem[Kolda and Bader, 2009]{kolda2009tensor}
Kolda, T.~G. and Bader, B.~W. (2009).
\newblock Tensor decompositions and applications.
\newblock {\em SIAM review}, 51(3):455--500.

\bibitem[Kozyrskiy and Phan, 2020]{kozyrskiy2020cnn}
Kozyrskiy, N. and Phan, A.-H. (2020).
\newblock Cnn acceleration by low-rank approximation with quantized factors.
\newblock {\em arXiv preprint arXiv:2006.08878}.

\bibitem[Krull et~al., 2019]{krull2019noise2void}
Krull, A., Buchholz, T.-O., and Jug, F. (2019).
\newblock Noise2void-learning denoising from single noisy images.
\newblock In {\em Proceedings of the IEEE/CVF Conference on Computer Vision and
  Pattern Recognition}, pages 2129--2137.

\bibitem[Lebedev et~al., 2014]{lebedev2014speeding}
Lebedev, V., Ganin, Y., Rakhuba, M., Oseledets, I., and Lempitsky, V. (2014).
\newblock Speeding-up convolutional neural networks using fine-tuned
  cp-decomposition.
\newblock {\em arXiv preprint arXiv:1412.6553}.

\bibitem[Lee et~al., 2019]{lee2019learning}
Lee, D., Kwon, S.~J., Kim, B., and Wei, G.-Y. (2019).
\newblock Learning low-rank approximation for cnns.
\newblock {\em arXiv preprint arXiv:1905.10145}.

\bibitem[Lehtinen et~al., 2018]{lehtinen2018noise2noise}
Lehtinen, J., Munkberg, J., Hasselgren, J., Laine, S., Karras, T., Aittala, M.,
  and Aila, T. (2018).
\newblock Noise2noise: Learning image restoration without clean data.
\newblock {\em Proc. 35th International Conference on Machine Learning}, pages
  2965--–2974.

\bibitem[Li et~al., 2020]{li2020understanding}
Li, J., Sun, Y., Su, J., Suzuki, T., and Huang, F. (2020).
\newblock Understanding generalization in deep learning via tensor methods.
\newblock In {\em International Conference on Artificial Intelligence and
  Statistics}, pages 504--515. PMLR.

\bibitem[Makitalo and Foi, 2012]{makitalo2012optimal}
Makitalo, M. and Foi, A. (2012).
\newblock Optimal inversion of the generalized anscombe transformation for
  poisson-gaussian noise.
\newblock {\em IEEE transactions on image processing}, 22(1):91--103.

\bibitem[Martin et~al., 2001]{MartinFTM01}
Martin, D., Fowlkes, C., Tal, D., and Malik, J. (2001).
\newblock A database of human segmented natural images and its application to
  evaluating segmentation algorithms and measuring ecological statistics.
\newblock In {\em Proc. 8th Int'l Conf. Computer Vision}, volume~2, pages
  416--423.

\bibitem[Nakajima et~al., 2011]{nakajima2011global}
Nakajima, S., Sugiyama, M., and Babacan, S. (2011).
\newblock Global solution of fully-observed variational bayesian matrix
  factorization is column-wise independent.
\newblock {\em Advances in Neural Information Processing Systems}, 24:208--216.

\bibitem[Nakajima et~al., 2013]{nakajima2013global}
Nakajima, S., Sugiyama, M., Babacan, S.~D., and Tomioka, R. (2013).
\newblock Global analytic solution of fully-observed variational bayesian
  matrix factorization.
\newblock {\em Journal of Machine Learning Research}, 14(Jan):1--37.

\bibitem[Novikov et~al., 2015]{novikov2015tensorizing}
Novikov, A., Podoprikhin, D., Osokin, A., and Vetrov, D.~P. (2015).
\newblock Tensorizing neural networks.
\newblock In {\em Advances in neural information processing systems}, pages
  442--450.

\bibitem[Phan et~al., 2020]{phan2020tensor}
Phan, A.-H., Cichocki, A., Uschmajew, A., Tichavsk{\`y}, P., Luta, G., and
  Mandic, D.~P. (2020).
\newblock Tensor networks for latent variable analysis: novel algorithms for
  tensor train approximation.
\newblock {\em IEEE transactions on neural networks and learning systems},
  31(11):4622--4636.

\bibitem[Quan et~al., 2020]{quan2020self2self}
Quan, Y., Chen, M., Pang, T., and Ji, H. (2020).
\newblock Self2self with dropout: Learning self-supervised denoising from
  single image.
\newblock In {\em Proceedings of the IEEE/CVF Conference on Computer Vision and
  Pattern Recognition}, pages 1890--1898.

\bibitem[Rabanser et~al., 2017]{rabanser2017introduction}
Rabanser, S., Shchur, O., and G{\"u}nnemann, S. (2017).
\newblock Introduction to tensor decompositions and their applications in
  machine learning.
\newblock {\em CoRR}, abs/1711.10781.

\bibitem[Richard and Montanari, 2014]{richard2014statistical}
Richard, E. and Montanari, A. (2014).
\newblock A statistical model for tensor {PCA}.
\newblock In {\em Advances in Neural Information Processing Systems}, pages
  2897--2905.

\bibitem[Ronneberger et~al., 2015]{ronneberger2015u}
Ronneberger, O., Fischer, P., and Brox, T. (2015).
\newblock U-net: Convolutional networks for biomedical image segmentation.
\newblock In {\em International Conference on Medical image computing and
  computer-assisted intervention}, pages 234--241. Springer.

\bibitem[Rumelhart et~al., 1986]{rumelhart1986learning}
Rumelhart, D.~E., Hinton, G.~E., and Williams, R.~J. (1986).
\newblock Learning representations by back-propagating errors.
\newblock {\em nature}, 323(6088):533--536.

\bibitem[Shi and Liu, 2023]{shi2023regularization}
Shi, B. and Liu, K. (2023).
\newblock Regularization by multiple dual frames for compressed sensing
  magnetic resonance imaging with convergence analysis.
\newblock {\em IEEE/CAA Journal of Automatica Sinica}.

\bibitem[Shi et~al., 2023]{shi2023provable}
Shi, B., Wang, Y., and Li, D. (2023).
\newblock Provable general bounded denoisers for snapshot compressive imaging
  with convergence guarantee.
\newblock {\em IEEE Transactions on Computational Imaging}, 9:55--69.

\bibitem[Soltanayev and Chun, 2018]{soltanayev2018training}
Soltanayev, S. and Chun, S.~Y. (2018).
\newblock Training deep learning based denoisers without ground truth data.
\newblock {\em Advances in neural information processing systems}, 31.

\bibitem[Tucker, 1966]{tucker1966some}
Tucker, L.~R. (1966).
\newblock Some mathematical notes on three-mode factor analysis.
\newblock {\em Psychometrika}, 31(3):279--311.

\bibitem[Ulyanov et~al., 2018]{ulyanov2018deep}
Ulyanov, D., Vedaldi, A., and Lempitsky, V. (2018).
\newblock Deep image prior.
\newblock In {\em Proceedings of the IEEE conference on computer vision and
  pattern recognition}, pages 9446--9454.

\bibitem[Venkatakrishnan et~al., 2013]{venkatakrishnan2013plug}
Venkatakrishnan, S.~V., Bouman, C.~A., and Wohlberg, B. (2013).
\newblock Plug-and-play priors for model based reconstruction.
\newblock In {\em 2013 IEEE Global Conference on Signal and Information
  Processing}, pages 945--948. IEEE.

\bibitem[Vulovi{\'c} et~al., 2013]{vulovic2013image}
Vulovi{\'c}, M., Ravelli, R.~B., van Vliet, L.~J., Koster, A.~J., Lazi{\'c},
  I., L{\"u}cken, U., Rullg{\aa}rd, H., {\"O}ktem, O., and Rieger, B. (2013).
\newblock Image formation modeling in cryo-electron microscopy.
\newblock {\em Journal of structural biology}, 183(1):19--32.

\bibitem[Wang et~al., 2020]{wang2020practical}
Wang, Y., Huang, H., Xu, Q., Liu, J., Liu, Y., and Wang, J. (2020).
\newblock Practical deep raw image denoising on mobile devices.
\newblock In {\em European Conference on Computer Vision}, pages 1--16.
  Springer.

\bibitem[Wu et~al., 2020]{wu2020unpaired}
Wu, X., Liu, M., Cao, Y., Ren, D., and Zuo, W. (2020).
\newblock Unpaired learning of deep image denoising.
\newblock In {\em European conference on computer vision}, pages 352--368.
  Springer.

\bibitem[Wu et~al., 2018]{wu2018fused}
Wu, Y., Tan, H., Li, Y., Zhang, J., and Chen, X. (2018).
\newblock A fused cp factorization method for incomplete tensors.
\newblock {\em IEEE transactions on neural networks and learning systems},
  30(3):751--764.

\bibitem[Zhang and Xia, 2018]{zhang2018tensor}
Zhang, A. and Xia, D. (2018).
\newblock Tensor {SVD}: Statistical and computational limits.
\newblock {\em IEEE Transactions on Information Theory}, 64(11):7311--7338.

\bibitem[Zhang et~al., 2020]{zhang2020denoising}
Zhang, C., Han, R., Zhang, A.~R., and Voyles, P.~M. (2020).
\newblock Denoising atomic resolution 4d scanning transmission electron
  microscopy data with tensor singular value decomposition.
\newblock {\em Ultramicroscopy}, page 113123.

\bibitem[Zhang et~al., 2017]{zhang2017beyond}
Zhang, K., Zuo, W., Chen, Y., Meng, D., and Zhang, L. (2017).
\newblock Beyond a gaussian denoiser: Residual learning of deep cnn for image
  denoising.
\newblock {\em IEEE transactions on image processing}, 26(7):3142--3155.

\bibitem[Zhang et~al., 2018]{zhang2018ffdnet}
Zhang, K., Zuo, W., and Zhang, L. (2018).
\newblock Ffdnet: Toward a fast and flexible solution for cnn-based image
  denoising.
\newblock {\em IEEE Transactions on Image Processing}, 27(9):4608--4622.

\bibitem[Zhang and Zuo, 2017]{zhang2017image}
Zhang, L. and Zuo, W. (2017).
\newblock Image restoration: From sparse and low-rank priors to deep priors
  [lecture notes].
\newblock {\em IEEE Signal Processing Magazine}, 34(5):172--179.

\bibitem[Zheng et~al., 2020]{zheng2020unsupervised}
Zheng, D., Tan, S.~H., Zhang, X., Shi, Z., Ma, K., and Bao, C. (2020).
\newblock An unsupervised deep learning approach for real-world image
  denoising.
\newblock In {\em International Conference on Learning Representations}.

\bibitem[Zheng et~al., 2017]{zheng2017motioncor2}
Zheng, S.~Q., Palovcak, E., Armache, J.-P., Verba, K.~A., Cheng, Y., and Agard,
  D.~A. (2017).
\newblock Motioncor2: anisotropic correction of beam-induced motion for
  improved cryo-electron microscopy.
\newblock {\em Nature methods}, 14(4):331--332.

\bibitem[Zhussip et~al., 2019]{zhussip2019extending}
Zhussip, M., Soltanayev, S., and Chun, S.~Y. (2019).
\newblock Extending stein's unbiased risk estimator to train deep denoisers
  with correlated pairs of noisy images.
\newblock {\em Advances in neural information processing systems}, 32.

\end{thebibliography}

\appendix

\newpage{} 
\begin{center}
{\Large{}{}{}Supplementary Material for ``Self-supervised Denoising via Low-rank Tensor Approximated  Convolutional Neural Network'' }{\Large\par}
\par\end{center}

\pagenumbering{arabic} 
\renewcommand*{\thepage}{S\arabic{page}}

\global\long\def\theequation{S\arabic{equation}}%
\setcounter{equation}{0}

\global\long\def\thesection{S\arabic{section}}%
\setcounter{section}{0}

\global\long\def\thetable{S\arabic{table}}%
\setcounter{table}{0}

\global\long\def\thetable{S\arabic{figure}}%
\setcounter{figure}{0}

\global\long\def\thetable{S\arabic{algorithm}}%
\setcounter{algorithm}{0}

\renewcommand{\thefigure}{S\arabic{figure}}
\renewcommand{\thetable}{S\arabic{table}}
\renewcommand{\thealgorithm}{S\arabic{algorithm}}

These supplementary materials collect the implementation details and additional experimental results of the proposed algorithm in the main content. 

\section{\MakeUppercase{implementation details}}

\subsection{Details of U-net Architecture}

We implement the U-net with a similar architecture to the one in \cite{lehtinen2018noise2noise}. Different from \cite{lehtinen2018noise2noise}, we substitute the first two convolutional layers with spatial width $3$ by a single convolutional layer with spatial width $11$ as suggested in \cite{bepler2020topaz}. All the convolutional layers are implemented with \texttt{pad='same'}, i.e., the output has the same shape as the input. Except for the last layer followed by linear activation, the other convolutions are coupled with a ReLU activation function; the pooling layers used in the U-net are the max-pooling downsampling with the stride $2$, realized by function \texttt{nn.MaxPool2d} in the PyTorch package. The up-sampling layers are the nearest-neighbor upsampling block, realized by function \texttt{nn.Upsample} in the PyTorch package.

\subsection{Details of Training}

In each of our experiments, each image is first normalized by subtracting its individual mean pixel intensity and dividing by its standard deviation. After finishing the denoising procedure described earlier, each denoised image is multiplied by its original standard deviation and added by its original mean.

For denoising of moderate size images (e.g. $256\times 256 \times 1$), image patches of $32\times 32\times 1$ with possible overlapping are randomly sampled from each image. Then, all these patches are randomly rotated by $90$, $180$, or $270$ degrees (to avoid interpolation artifacts) and mirrored as data augmentation. After that, mini-batches of size 128 are formed to fuel the subsequent stochastic gradient descent (SGD) algorithm. There are around $m=60$ iterations in each epoch when denoising images of size $256\times 256 \times 1$.

For denoising of large size images (e.g. $5760 \times 4092\times 1$), image patches of $800\times 800\times 1$ are randomly selected with proper data augmentations. For the sake of computational capacity, mini-batches of size $4$ are formed for the optimizer. There are around $m=1000$ iterations in each epoch when denoising the micrographs of size $5760\times 4092 \times 1$. 
In addition, to avoid the edge artifacts, we include a padding of $500$ pixels when denoising the image by patches \citep{bepler2020topaz}.

\subsection{Details of Additional Algorithms}

Let $\mathcal{M}_k(\cdot)$ be the matricization operator of tensor ``$\cdot$" along its $k$-th mode (see \cite[Section 2.4]{kolda2001orthogonal} for a formal definition); $\mathcal{N}_d(\cdot;\boldsymbol{\mu},\Sigma)$ denotes the density of the $d$-dimensional Gaussian random variable with mean $\boldsymbol{\mu}$ and covariance matrix $\Sigma$; $\text{SVD}(A, r)$ is defined as the matrix comprised of the top $r$ left singular vectors of $A$. The implementations of VBMF (variational Bayes matrix factorization) and Partial HOOI (partial high-order orthogonal iteration), two key sub-algorithms in our proposed method, are provided in Algorithms \ref{al:VBMF} and \ref{al:PHOOI}, respectively. 
\begin{algorithm*}
\caption{Rank selection by variational Bayes matrix factorization (VBMF)
\label{al:VBMF}}
\begin{algorithmic}
\State
{\bfseries Input:} Weight kernel $\mathcal{W}\in \mathbb{R}^{d\times d \times s\times \widehat{s}}$, $\sigma^2 \in \mathbb{R}$
\State $M = \mathcal{M}_3(\mathcal{W})\in \mathbb{R}^{s\times (d^2\widehat{s})}$  and assume: $M = B A^\intercal$ with $B \in \mathbb{R}^{s \times h}, A\in \mathbb{R}^{(d\times d\times \widehat{s})\times h}$
\State $M = \sum_{i=1}^h \gamma_i \boldsymbol{w}_{bi}\boldsymbol{w}_{ai}^\intercal$ with $\gamma_1\geq \cdots \geq \gamma_h$
\State
\Comment{Gaussian Priors on $A=(\boldsymbol{a}_1,\cdots,\boldsymbol{a}_h)$ and $B=(\boldsymbol{b}_1,\cdots,\boldsymbol{b}_h)$ }
$$\phi_{\mathrm{A}}(A) \propto \exp \left\{-\sum_{i=1}^{h} {\left\|\boldsymbol{a}_{i}\right\|^{2}}/{(2 c_{a_{i}}^{2})}\right\} \text { and } \phi_{\mathrm{B}}(B) \propto \exp \left\{-\sum_{i=1}^{h} {\left\|\boldsymbol{b}_{i}\right\|^{2}}/{(2 c_{b_{i}}^{2})}\right\}$$
\State
$r(A,B\mid \mathcal{W}) = \prod_{i=1}^h
\left\{\mathcal{N}_{d\times d\times \widehat{s}}(\boldsymbol{a}_i;\boldsymbol{\mu}_{ai},\Sigma_{ai})\times
\mathcal{N}_{s}(\boldsymbol{b}_i;\boldsymbol{\mu}_{bi},\Sigma_{bi})\right\}$
\hfill\Comment{Assume probabilistic independence of $A$ and $B$ given $\mathcal{W}$}
\State $F_{\text{VB}}(r\mid \mathcal{W})=
F_{\text{VB}}(\{\boldsymbol{\mu}_{ai},\boldsymbol{\mu}_{bi}, \Sigma_{ai},\Sigma_{bi}:i=1,\cdots,h\})$
\hfill\Comment{The variational Bayes free energy $F_{\text{VB}}$ can be analytically derived}
\State $\{\boldsymbol{\mu}^*_{ai},\boldsymbol{\mu}^*_{bi}, \Sigma^*_{ai},\Sigma^*_{bi}\}=\arg\min F_{\text{VB}}(r\mid \mathcal{W})$
\hfill\Comment{The global solution can then be analytically obtained}
\State $\alpha_i^* = \|\boldsymbol{\mu}_{ai}\|^2+\text{tr}(\Sigma_{ai}^*),
\beta_i^* = \|\boldsymbol{\mu}_{bi}\|^2+\text{tr}(\Sigma_{bi}^*)$
\State $c_{a_i}^{2*}=\alpha_i^*/(d\times d\times \widehat{s}),c_{b_i}^{2*}=\beta^*_i/s$
\State $\widetilde{\gamma}_{i}=\sqrt{\frac{(2d+s+\widehat{s}) \sigma^{2}}{2 }+\frac{\sigma^{4}}{2c_{a_{i}}^{2} c_{b_{i}}^{2*}}+\sqrt{\left(\frac{(2d+s+\widehat{s}) \sigma^{2}}{2}+\frac{\sigma^{4}}{ c_{a_{i}}^{2*} c_{b_{i}}^{2*}}\right)^{2}-{2ds\widehat{s} \sigma^{4}}}}$
\hfill\Comment{Define the cutoff value}
\State {\bfseries Output:} Global optimal mode-3 rank $r_3= \arg \max_i
\{\gamma_i>\Tilde{\gamma}_i: i=1,\cdots,h\}$ of $\mathcal{W}$ (optimal mode-4 rank $r_4$ can be deduced analogously)
\end{algorithmic}
\end{algorithm*}

\begin{algorithm*}
\caption{Partial higher order orthogonal iteration (Partial HOOI) for partial Tucker decomposition \label{al:PHOOI}}
\begin{algorithmic}
\State
{\bfseries Input:} Weight kernel $\mathcal{W}\in \mathbb{R}^{d\times d \times s\times \widehat{s}}$, mode-3 and mode-4 ranks $(r_3,r_4)$
\State 
$U_{r_3}^{(0)}=\text{SVD}\{\mathcal{M}_3(\mathcal{W}),r_3\},
U_{r_4}^{(0)}=\text{SVD}\{\mathcal{M}_4(\mathcal{W}),r_4\}$ \hfill  \Comment{Initialization}
 \For{$k=0,1,\cdots, K-1$}
  \State $\mathcal{G}^{(k)} = 
  \mathcal{W} \times_3 (U_{r_3}^{(k)})^\intercal 
  \times_4
  (U_{r_4}^{(k)})^\intercal$
  \State $U^{(k+1)}_{r_3} = \text{SVD}\{\mathcal{M}_3(\mathcal{\mathcal{G}}^{(k)}),r_3\}$
  \State $U^{(k+1)}_{r_4} = \text{SVD}\{\mathcal{M}_4(\mathcal{\mathcal{G}}^{(k)}),r_4\}$
 \EndFor
\State $\mathcal{G}^{(K)} = 
\mathcal{W} \times_3 (U_{r_3}^{(K)})^\intercal 
\times_4
(U_{r_4}^{(K)})^\intercal
$
\State {\bfseries Output:} Decomposed parts $\mathcal{G}, U_{r_3}, U_{r_4}$
\end{algorithmic}
\end{algorithm*}

\section{\MakeUppercase{Additional Numerical Experiments}}
\label{sec:additional_exps}

For completeness, we compare our method with other existing representative self-supervised learning methods in Table \ref{tab:more_methods} in terms of their in-sample performance, i.e., trained on each image individually. 
\begin{table}[htbp]
    \caption{In-sample performance in terms of PSNR/SSIM for synthetic additive Gaussian noise removal}
    \vspace{0.15cm}
    \centering
\resizebox{\columnwidth}{!}{%
    \begin{tabular}{cccccc}
    \toprule
      & N2N \citep{lehtinen2018noise2noise}  & N2V \citep{krull2019noise2void} & Neighbor2Neighbor \citep{huang2021neighbor2neighbor} &NN+BM3D &NN+BM3D+T  \\
     \midrule
    SET12& 30.66dB/0.95& 28.84dB/0.80 & \textbf{31.09dB/0.86}&29.81dB/0.89& 30.41dB/0.91 \\
    BSD68 &  28.86dB/0.82 &27.72dB/0.79 &\textbf{30.79dB/0.87}&27.42dB/0.78&28.80dB/0.80\\
    \bottomrule
    \end{tabular}
    }
    \label{tab:more_methods}
\end{table}

N2V is trained on unorganized images and N2N is trained on paired noisy images. Intuitively, they are expected to outperform our method since they have more information available during training. However, our method only subject to small PSNR/SSIM drops and it can achieve comparable state-of-the-art performance as the Neighbor2Neighbor, which requires computational subsampling scheme.

\begin{figure*}[htbp]
    \centering
    \includegraphics[width=\linewidth]{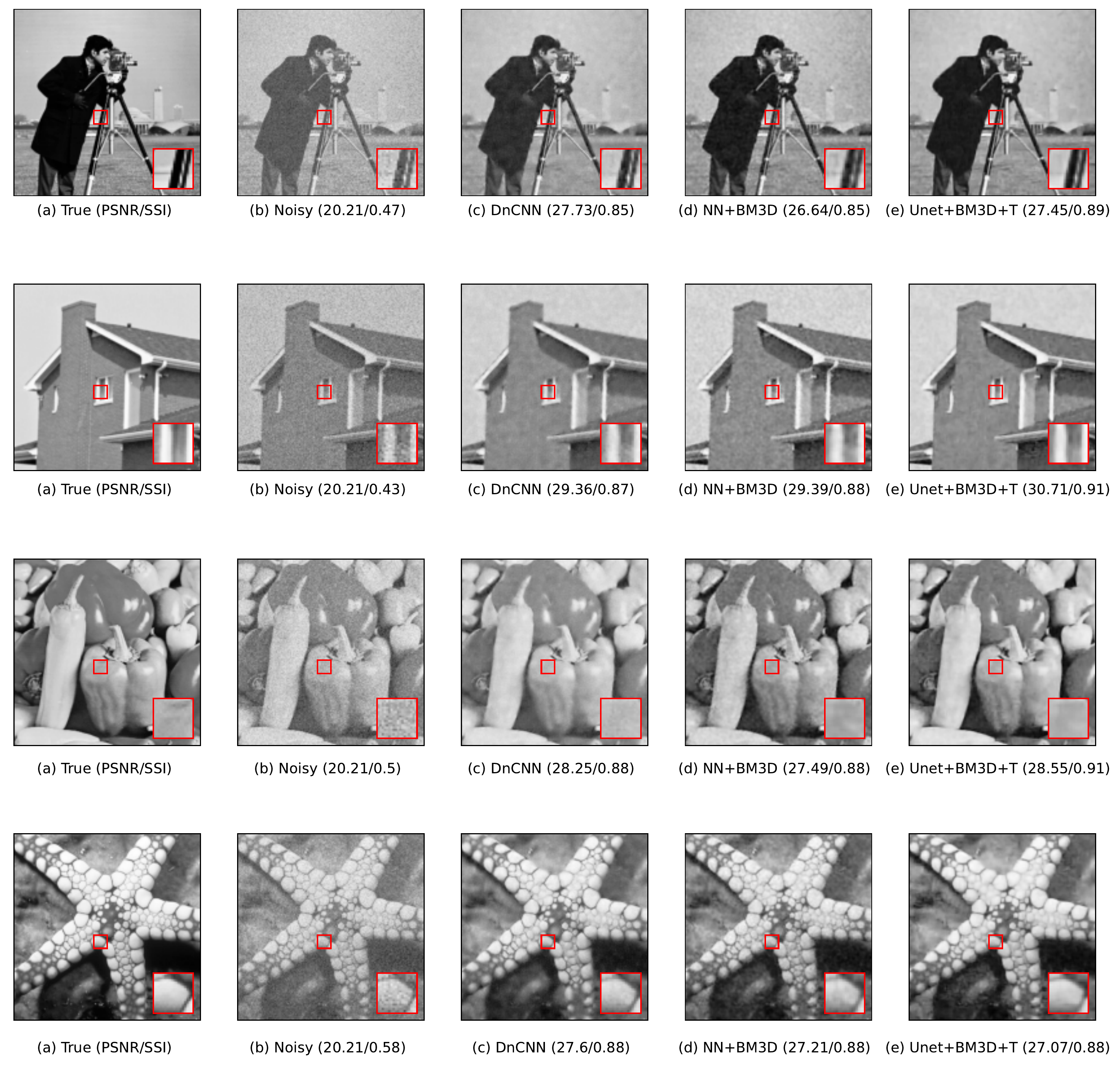}
    \caption{Visualization of out-sample denoising results with PNSR/SSIM on the (rest) $\textit{SET12}$. Noted that our denoiser is trained based on \textit{LENA}.}
    \label{fig:set12:1}
\end{figure*}

\begin{figure*}[htbp]
    \centering
    \includegraphics[width=\linewidth]{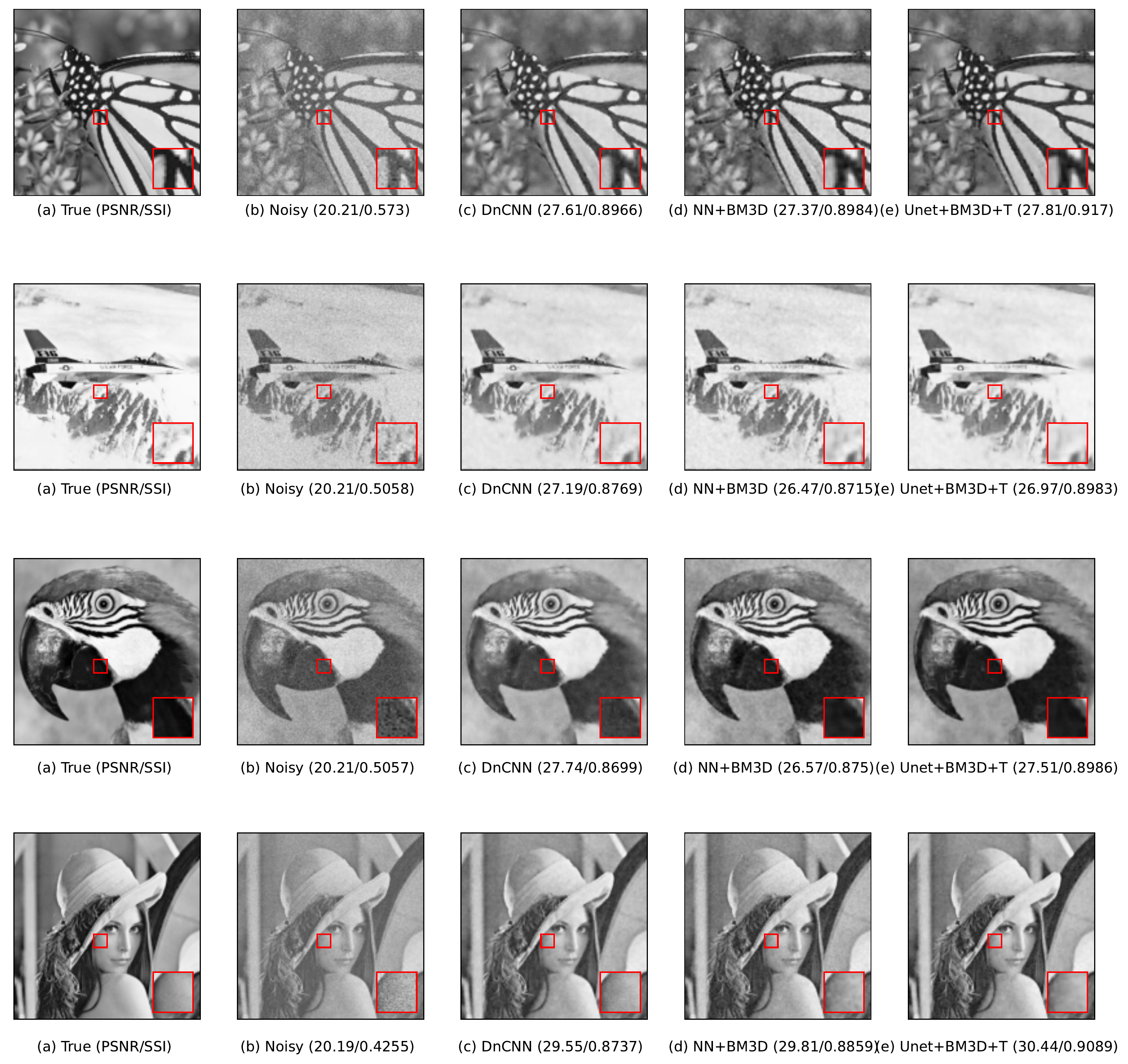}
    \caption{Visualization of out-sample denoising results with PNSR/SSIM on the (rest) $\textit{SET12}$. Noted that our denoiser is trained based on \textit{LENA}.}
    \label{fig:set12:2}
\end{figure*}

\begin{figure*}[htbp]
    \centering
    \includegraphics[width=\linewidth]{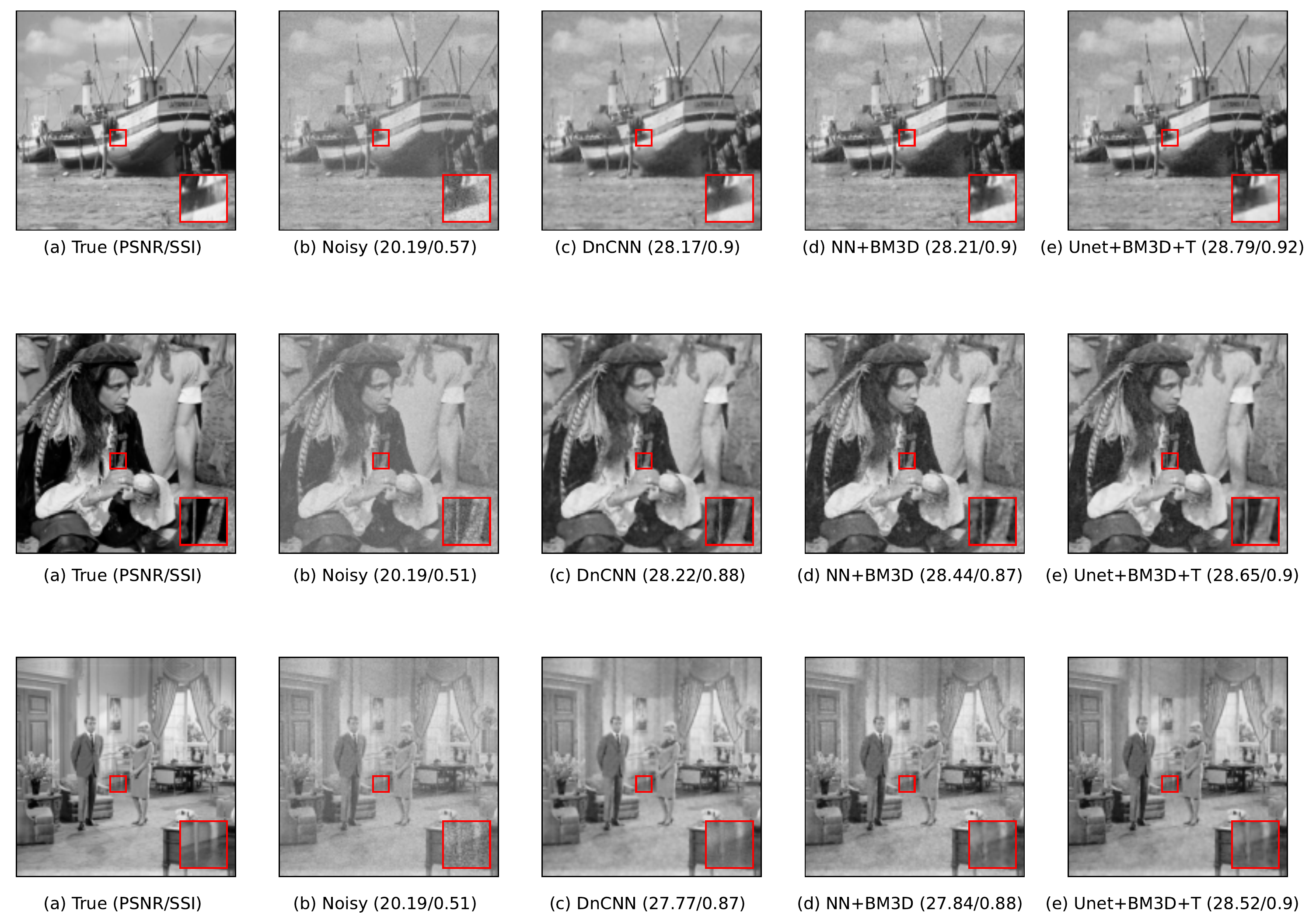}
    \caption{Visualization of out-sample denoising results with PNSR/SSIM on the (rest) $\textit{SET12}$. Noted that our denoiser is trained based on \textit{LENA}.}
    \label{fig:set12:3}
\end{figure*}

\begin{figure*}[htbp]
    \centering
    \includegraphics[width=\linewidth]{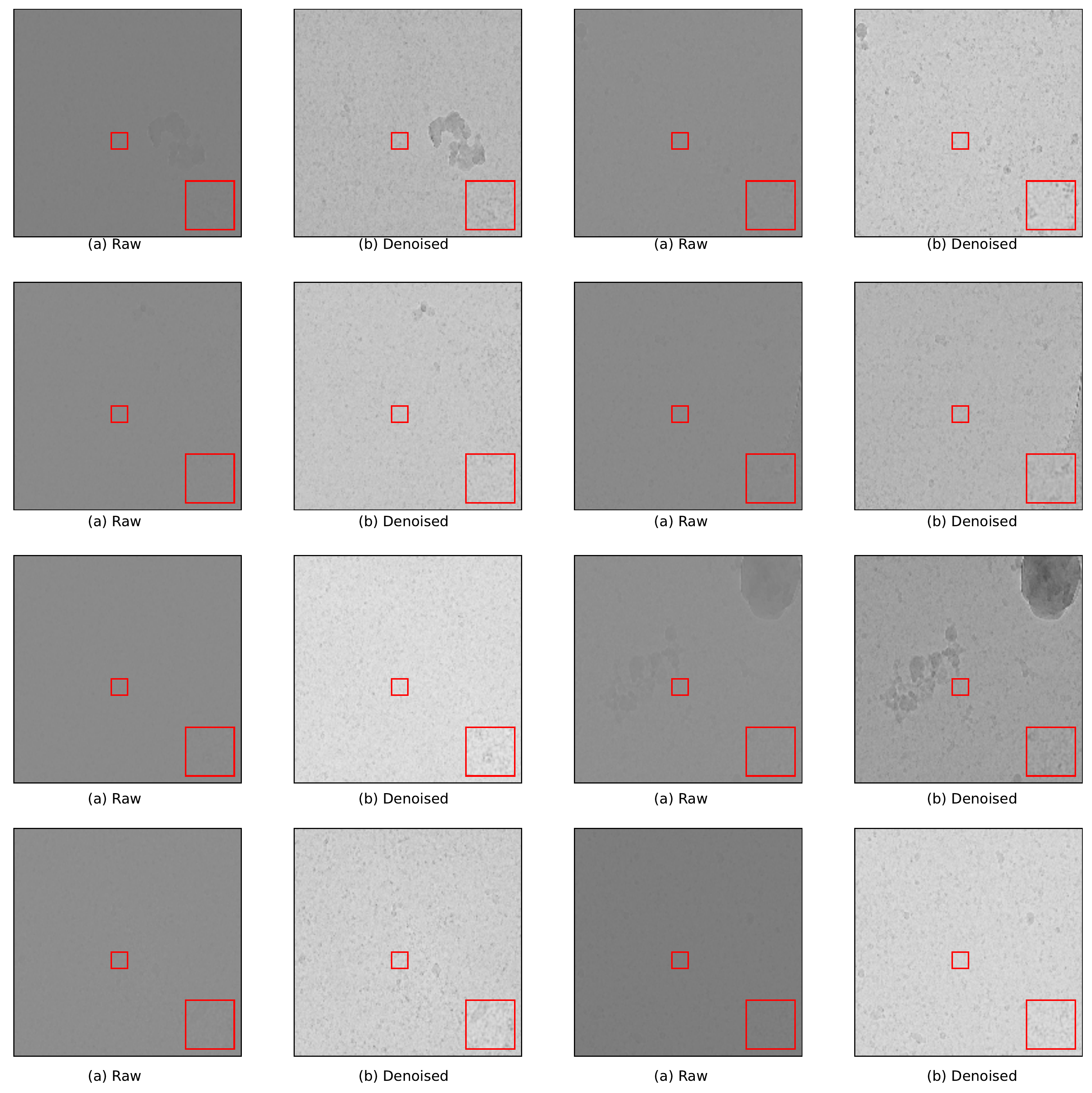}
    \caption{Visualization of out-sample denoising results on the Cryo-EM images of SARS-CoV-2 2P protein with the proposed algorithm coupled with VST.}
    \label{fig:real:more}
\end{figure*}

Figures \ref{fig:set12:1}, \ref{fig:set12:2}, and \ref{fig:set12:3} visualize the out-sample denoising performance of our denoiser trained on $\textit{LENA}$ on the other images in $\textit{SET12}$. Figures \ref{fig:S1}--\ref{fig:S17} provide more experimental results on the generalization ability of our trained model on the external datasets $\textit{BSD68}$. In Figure \ref{fig:real:more}, we provide more denoising results for real Cryo-EM images on SARS-CoV-2 2P protein.


\foreach \x in {1,2,...,17}
{
    \begin{figure*}[htbp]
        \includegraphics[width=\linewidth]{figures/model_performance_test_others\x.pdf}
        \caption{Visualization of out-sample denoising results with PNSR/SSIM on the external image sets $\textit{BSD68}$. Noted that our denoiser is trained based on \textit{LENA}.}
        \label{fig:S\x}
    \end{figure*}
}

\end{document}